\documentclass[10pt,twocolumn,letterpaper]{article}

\usepackage{cvpr}              

\usepackage{multirow}
\usepackage{booktabs}
\usepackage{amssymb}
\usepackage{amsmath}
\usepackage{xcolor}
\usepackage{graphicx} 
\usepackage{algorithm}
\usepackage{algorithmic}
\usepackage{newfloat}
\usepackage{listings}
\usepackage{subcaption}
\usepackage{array}
\usepackage{pifont}
\usepackage{colortbl}
\usepackage[most]{tcolorbox}
\usepackage{stfloats}
\usepackage{placeins}

\definecolor{cvprblue}{rgb}{0.21,0.49,0.74}
\usepackage[pagebackref,breaklinks,colorlinks,allcolors=cvprblue]{hyperref}

\def\paperID{5631} 
\def\confName{CVPR}
\def\confYear{2026}

\title{From Intuition to Investigation: A Tool-Augmented Reasoning MLLM Framework for Generalizable Face Anti-Spoofing}

\author{
    Haoyuan Zhang\textsuperscript{\rm1,\rm2,\rm3 \dag},
    Keyao Wang\textsuperscript{\rm3 }\setcounter{footnote}{1}\thanks{Equal contribution; \textsuperscript{*}Corresponding author}\ \ ,
    Guosheng Zhang\textsuperscript{\rm3},
    Haixiao Yue\textsuperscript{\rm3}, \\
    Zhiwen Tan\textsuperscript{\rm3},
    Siran Peng\textsuperscript{\rm1,\rm2},
    Tianshuo Zhang\textsuperscript{\rm1,\rm2},
    Xiao Tan\textsuperscript{3},
    Kunbin Chen\textsuperscript{\rm3},
    Wei He\textsuperscript{3}, \\
    Jingdong Wang\textsuperscript{\rm3},
    Ajian Liu\textsuperscript{\rm1,\rm2},
    Xiangyu Zhu\textsuperscript{\rm1,\rm2},
    Zhen Lei\textsuperscript{\rm1,\rm2,\rm4,\rm5 $\ast$} \and
    \textsuperscript{\rm1}SAI, UCAS;
    \textsuperscript{\rm2}MAIS, CASIA;
    \textsuperscript{\rm3}Baidu Inc;
    \textsuperscript{\rm4}CAIR, HKISI, CAS; \textsuperscript{\rm5}M.U.S.T \\
    {\tt\small \{zhanghaoyuan2023,pengsiran2023,ajian.liu,xiangyu.zhu,zhen.lei\}@ia.ac.cn} \\
    {\tt\small \{wangkeyao,zhangguosheng,yuehaixiao,tanzhiwen,tanxiao01,chenkunbin,hewei06\}@baidu.com} \\
    {\tt\small tianshuo.zhang@nlpr.ia.ac.cn,wangjingdong@outlook.com}
}
\begin{document}
\maketitle
\begin{abstract}
Face recognition remains vulnerable to presentation attacks, calling for robust Face Anti-Spoofing (FAS) solutions. Recent MLLM-based FAS methods reformulate the binary classification task as the generation of brief textual descriptions to improve cross-domain generalization. However, their generalizability is still limited, as such descriptions mainly capture intuitive semantic cues (e.g., mask contours) while struggling to perceive fine-grained visual patterns. To address this limitation, we incorporate external visual tools into MLLMs to encourage deeper investigation of subtle spoof clues. Specifically, we propose the Tool-Augmented Reasoning FAS (TAR-FAS) framework, which reformulates the FAS task as a Chain-of-Thought with Visual Tools (CoT-VT) paradigm, allowing MLLMs to begin with intuitive observations and adaptively invoke external visual tools for fine-grained investigation. To this end, we design a tool-augmented data annotation pipeline and construct the ToolFAS-16K dataset, which contains multi-turn tool-use reasoning trajectories. Furthermore, we introduce a tool-aware FAS training pipeline, where Diverse-Tool Group Relative Policy Optimization (DT-GRPO) enables the model to autonomously learn efficient tool use. Extensive experiments under a challenging one-to-eleven cross-domain protocol demonstrate that TAR-FAS achieves SOTA performance while providing fine-grained visual investigation for trustworthy spoof detection.
\end{abstract}
\vspace{-0.2cm}
\section{Introduction}
\renewcommand{\thefootnote}{}
\footnotetext{Work done during Haoyuan Zhang's internship at Baidu Inc.}
\label{sec:intro}
\vspace{-0.1cm}

Rapid advances in facial recognition system have enabled convenient, contactless use in payments and identity verification. Nevertheless, such systems are still susceptible to various external influences and spoofing attacks, including printed photographs~\cite{casiafasd}, replayed videos~\cite{oulunpu}, and even realistic 3D masks~\cite{hifimask}. To address these vulnerabilities, Face Anti-Spoofing (FAS) techniques have been introduced to strengthen the reliability of facial recognition. Early FAS approaches~\cite{siw,cnn2} have achieved strong performance in intra-domain settings, but their generalization significantly degrades when deployed in cross-domain environments involving unseen conditions or spoofing materials.

\begin{figure}
    \centering
   \includegraphics[width=\linewidth]{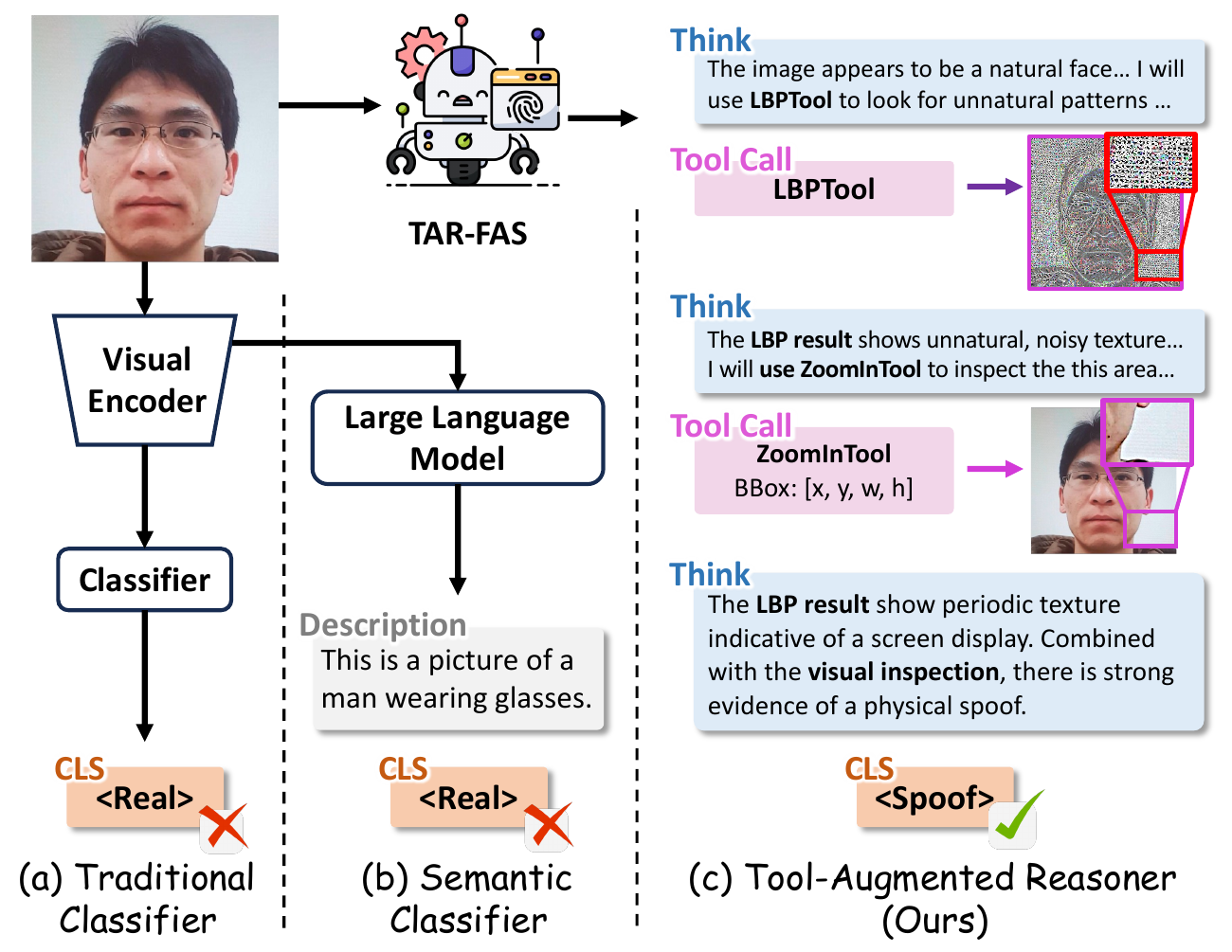}
   \caption{Previous methods struggle to distinguish high-quality spoof samples with subtle visual cues, while TAR-FAS enhances fine-grained investigation by coupling MLLM with visual tools.}
   \label{fig:front}
   \vspace{-0.4cm}
\end{figure}

To improve cross-domain generalization in FAS, some prior studies~\cite{sda, liu2022source, GDA, cdftn, liu2023visual} reduce the domain gap by aligning features with access to target data. Another line of research~\cite{iadg, cai2025towards, liu2024moeit, gacfas, ttdg} seeks to learn domain-invariant representations using adversarial learning or feature disentanglement. However, large domain discrepancies and varied spoofing types still hinder robust generalization. Recently, CLIP-based frameworks~\cite{flip, cfpl, scptl, tffas} have been introduced to enhance generalizability by aligning images with manually designed class-wise descriptions (e.g., \textit{This is a \textbf{real / spoof} face.}). Going a step further, I-FAS~\cite{ifas} introduces sample-wise descriptions and develops a pioneering MLLM framework that generates both classification results and one-sentence explanations.

However, researchers have found that MLLMs often exhibit a certain degree of blindness to low-level visual features~\cite{lowlevel_blind1, lowlevel_blind2}. The use of brief descriptions in existing MLLM-based FAS methods \cite{ifas} further amplifies this tendency, encouraging the model to focus on coarse semantic cues (e.g., screen borders or mask contours) while neglecting fine-grained spoof traces. As a result, these methods may miss subtle visual evidence that is important for reliable FAS, which reduces their generalization ability. This raises an important question: how can we guide MLLMs to perceive subtle spoof cues that are easily overlooked? Fortunately, MLLMs have an inherent ability to call external tools during the Chain-of-Thought (CoT) reasoning process, providing a possible way to overcome this limitation. In the history of FAS, traditional methods have often employed basic visual operators (e.g., LBP~\cite{lbp2, lbptop}, HOG~\cite{hog}, and FFT~\cite{fft1, fft2}) to extract fine-grained features and enhance generalization. Inspired by this idea, we incorporate external visual tools into MLLM-based FAS framework to encourage the model toward deeper exploration of hidden visual clues. This enables the model to move beyond coarse intuition to fine-grained investigation.

In this work, we propose Tool-Augmented Reasoning FAS (TAR-FAS), a framework that drives MLLMs from intuition to investigation by enabling them to adaptively invoke appropriate visual tools during the reasoning process. Specifically, TAR-FAS reformulates FAS as a Chain-of-Thought with Visual Tools (CoT-VT) paradigm as shown in \Cref{fig:front}, where the model can capture fine-grained visual cues via external visual tools. To equip the MLLM with this capability, we design a tool-augmented data annotation pipeline and construct the ToolFAS-16K dataset, which contains multi-turn tool-use reasoning trajectories. Within this pipeline, we introduce an expert-model-guided mechanism to ensure annotation reliability. Leveraging ToolFAS-16K along with fundamental FAS datasets, we further develop a tool-aware FAS training pipeline that equips the model with effective tool-based reasoning capability. We use ToolFAS-16K dataset to inject the tool-call format into the model’s instruction space, and introduce Diverse-Tool Group Relative Policy Optimization (DT-GRPO) which encourages the model to utilize different tools through a tool-diversity reward function, thus enables it to autonomously learn efficient and adaptive tool use solely from query–label pairs. Under the challenging one-to-eleven cross-domain testing protocol, TAR-FAS achieves state-of-the-art (SOTA) performance and produces interpretable reasoning chains that demonstrate a clear transition from coarse intuition to fine-grained investigation, validating both the effectiveness of the proposed training pipeline and the constructed ToolFAS-16K dataset.
Our contributions can be summarized as follows:
\begin{itemize}
    \item We are the first to reformulates FAS as a Chain-of-Thought with Visual Tools (CoT-VT) paradigm and propose Tool-Augmented Reasoning FAS (TAR-FAS) framework, enabling MLLMs to adaptively invoke external visual tools for robust spoof detection.
    \item We introduce a tool-augmented data annotation pipeline, and construct ToolFAS-16K dataset containing multi-turn tool-use reasoning trajectories. We further introduce a tool-aware FAS training pipeline containing a Diverse-Tool Group Relative Policy Optimization (DT-GRPO) for autonomous tool-use learning.
    \item Extensive experiments on the challenging cross-domain benchmarks (One-to-Eleven) demonstrate that our method achieves a significant improvement over state-of-the-art (SOTA) methods. 
\end{itemize}
\section{Related Works}
\label{sec:relatedwork}

\subsection{Face Anti-Spoofing}

FAS aims to distinguish real live faces from presentation attacks such as printed photos, replay videos, or 3D masks. Early works primarily relied on handcrafted features, which later evolved into deep learning–based methods \cite{patchnet,zhang2020face,wang2024csdg}. With the rise of deep networks, CNN-based~\cite{yu2020searching} and Transformer-based approaches~\cite{vitfas} have focused on designing task-specific architectures for robust representation learning. In addition, auxiliary supervision (e.g., depth maps~\cite{siw} and reflection maps~\cite{zhang2021structure}) has been widely employed to provide additional cues. Despite achieving impressive intra-dataset results, these methods often fail to generalize under domain shifts. To improve cross-domain robustness, Domain Adaptation (DA) techniques~\cite{sda, liu2022source, GDA, cdftn, liu2023visual} attempt to reduce the distribution gap between source and target domains using unlabeled target data, while Domain Generalization (DG) approaches~\cite{iadg, cai2025towards, liu2024moeit, gacfas, ttdg} aim to learn domain-invariant representations across multiple sources through adversarial learning~\cite{maddg, ssdgr} or meta-learning~\cite{hfnmp, d2am}. 
However, most existing approaches rely solely on binary supervision to extract generalized liveness features, limiting their capacity to model complex cross-domain variations.

\subsection{Multimodal Large Language Models}

Early Vision-Language Models (VLMs)~\cite{clip, huang2024empirical, jiang2024effectiveness} focused on learning aligned vision–language embeddings, while instruction-tuned models such as BLIP-2~\cite{li2023blip} and LLaVA~\cite{liu2023visual} established the MLLM paradigm by connecting vision encoders with frozen LLMs through cross-modal connectors. More recent efforts like Qwen-VL~\cite{qwen25vl} and InternVL~\cite{internvl3} further improve spatial grounding and fine-grained perception. In FAS, FLIP \cite{flip} and CFPL-FAS \cite{cfpl} utilized the pretrained CLIP \cite{clip} model to align visual features with language supervision for better generalization. I-FAS~\cite{ifas} extended this paradigm by framing FAS as a visual question answering task, bridging perception and reasoning for enhanced generalization and interpretability. Nevertheless, performance remains limited, as simple textual descriptions capture only coarse semantic cues but struggle to model fine-grained visual patterns.

\subsection{Tool-Use MLLM Agent}

Recent advances extend MLLMs from static perception to dynamic tool-use agents capable of invoking external functions for enhanced reasoning and control. Early frameworks such as ReAct and Chameleon~\cite{react, chameleon} demonstrated how language models can couple reasoning with external actions like API calls and retrieval. Recent efforts, including ``thinking with images''~\cite{thinkwithimages} and DeepEyes~\cite{deepeyes}, further expand this paradigm to the vision–language domain, enabling stepwise visual tool interactions for complex reasoning. Building on these advances, we make the first attempt to integrate visual tools with MLLM into FAS task for robust detection beyond intuitive classification.
\section{Methodology}

\subsection{Tool-Augmented Data Annotation}

The pipeline include data and tool selection, expert-model-guided annotation workflow, and data verification. Through this pipeline, we construct a multi-turn tool-augmented ToolFAS-16K dataset with the help of Gemini-2.5 pro.

\subsubsection{Data and Tool Selection}

\begin{figure}[t]
\centering
\includegraphics[width=\linewidth]{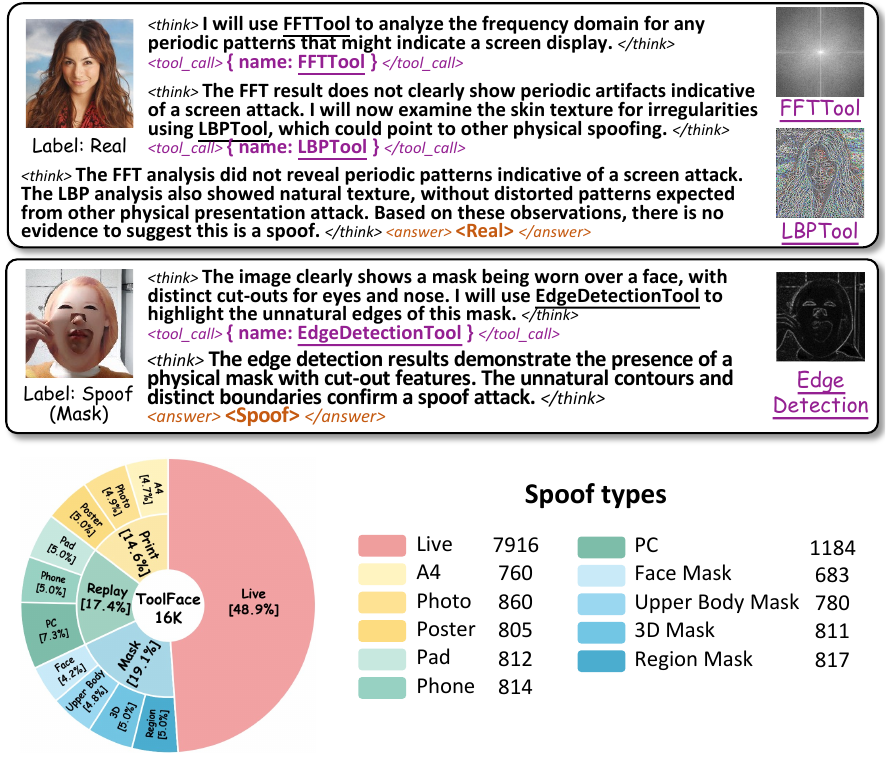}
\caption{Detailed construction and samples of ToolFAS-16K.}
\label{fig:data}
\vspace{-0.3cm}
\end{figure}

\noindent \textbf{Data Selection.}
CelebA-Spoof \cite{celebaspoof} is a large-scale, real-world face anti-spoofing dataset with various attack types. We select a total of 16,172 images from CelebA-Spoof, covering real samples and 10 different spoof types. The detailed data sample and distribution is illustrated in \Cref{fig:data}.

\noindent \textbf{Tool Selection.}
We select a set of visual tools based on their proven effectiveness in prior FAS research, aiming to extract low-level spatial and frequency-domain cues:
\begin{itemize}
    \item \textbf{Zoom-In Tool.} This tool takes a bounding box input to focus on facial regions for fine-grained local analysis.
    \item \textbf{Texture-based Tools.} The LBP operator captures subtle texture irregularities, which are effective for identifying mask materials and paper surface inconsistencies~\cite{lbp2, lbptop}.
    \item \textbf{Frequency-based Tools.} The FFT and Wavelet Transform operators analyze frequency domain representations, revealing abnormal spectral distributions typically caused by display and print attacks~\cite{fft1, fft2, wavelet1, wavelet2}.  
    \item \textbf{Structure-based Tools.} The Laplacian Edge Detection and HOG operators emphasize structural boundaries and reflectance discontinuities, helping to expose mask edges and screen boundary artifacts~\cite{laplacian1, laplacian2}. 
\end{itemize}
This combination provides diverse and complementary visual evidence that enhances the robustness of our tool-augmented reasoning framework.

\subsubsection{Annotation Workflow}

\begin{figure}[t]
   \centering
   \includegraphics[width=\linewidth]{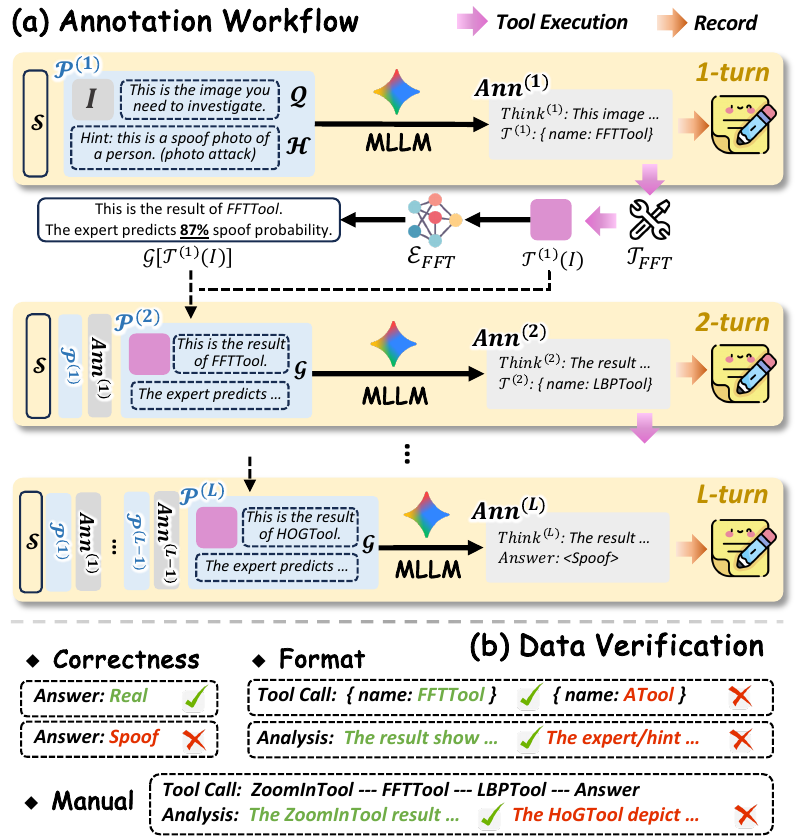}
   \caption{Illustration of the data annotation pipeline. (a) Each annotation can be divided into multiple sub-annotations $Ann^{(l)}$ generated by the MLLM with historical context, and the multi-turn process ends when the MLLM outputs the final answer. (b) The data verification process ensuring annotation reliability.}
   \label{fig:annotation}
\vspace{-0.4cm}
\end{figure}

The detailed annotation workflow is illustrated in \Cref{fig:annotation}-(a). Specifically, for each sample , we annotate through a multi-turn process and get the annotation $Ann$ from multi-turn sub-annotation $Ann^{(l)}$ concatenation:
\begin{equation}
    Ann = [Ann^{(1)}; Ann^{(2)}; \cdots ; Ann^{(L)}],
\end{equation}
where $L$ is the turn number of each annotation which will not exceed a max number of $L^{max}$ (we set $L^{max}=6$ in this paper). In each turn, sub-annotation is generated by a MLLM with historical context as input:
\begin{equation}
    Ann^{(l)} = MLLM(\mathcal{S}, \mathcal{P}^{(<l)}, Ann^{(<l)}, \mathcal{P}^{(l)}),
\end{equation}
where $\mathcal{S}$ denotes the system prompt (detail in supplementary materials), $P^{(l)}$ denotes the user prompt for turn $l$. 

Each sub-annotation generated by MLLM is constrained using structured output of Gemini API which can be parsed into reasoning-tool or reasoning-answer pair:
\begin{equation}
Ann^{(l)} =
\begin{cases}
\text{Think}^{(l)} + \mathcal{T}^{(l)} \\
\text{Think}^{(l)} + \text{Answer}
\end{cases},
\end{equation}
where $\mathcal{T}^{(l)} \in \{ \mathcal{T}_1, \mathcal{T}_2, \cdots, \mathcal{T}_K \}$ denotes the tool call of $l$ turn, $K$ denotes the tool number. We will end the multi-turn annotation process if the MLLM give the final answer.

In the first turn, user prompt $\mathcal{P}^{(1)}$ contains a input image $\mathcal{I}$, a textual query $\mathcal{Q}$ (``\textit{This is the image you need to investigate.}'') and a hint description $\mathcal{H}(\mathcal{I})$ (e.g. photo attack, phone attack) of the input image:
\begin{equation}
    \mathcal{P}^{(1)} = \mathcal{I} \oplus \mathcal{Q} \oplus \mathcal{H}(\mathcal{I}),
\end{equation}
where $\oplus$ denotes concatenation.

After the first round, user prompt $\mathcal{P}^{(l)}$ for each turn includes tool execution results and expert guidance:
\begin{equation}\label{eqn:guidance}
\mathcal{P}^{(l)} = \mathcal{T}^{(l-1)}(\mathcal{I}) \oplus \mathcal{G}[\mathcal{T}^{(l-1)}(\mathcal{I})], \,\, 1 < l \leq L \,\,\,,
\end{equation}
where $\mathcal{T}^{(l-1)}$ denotes the tool-call request parsed from $l-1$ turn sub-annotation $Ann^{(l-1)}$, $\mathcal{G}(\cdot)$ denotes the tool-result guidance of expert-model-guided mechanism.

\begin{figure*}[t]
   \centering
   \includegraphics[width=\linewidth]{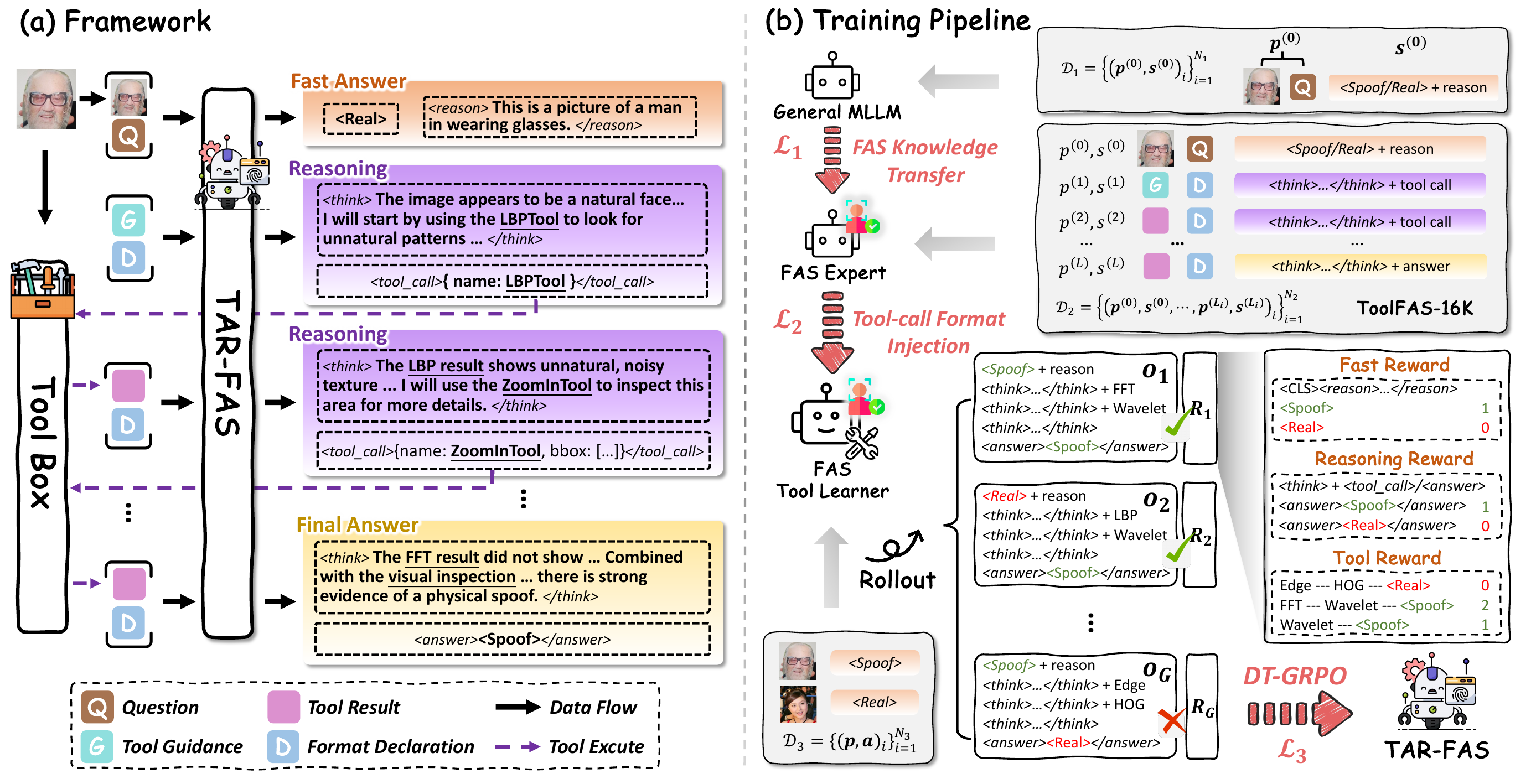}
   \caption{The total framework and training pipeline of TAR-FAS. (a) introduce the framework of TAR-FAS which can give a quick intuitive answer and investigate with visual tools to give a more accurate final decision. (b) illustrate the whole tool-aware FAS training pipeline including FAS knowledge transfer, tool-call format injection and DT-GRPO.}
   \label{fig:method}
\vspace{-0.4cm}
\end{figure*}

\noindent \textbf{Expert-Model-Guided Mechanism.}
We leverage multiple lightweight expert networks to generate tool-result guidance $\mathcal{G}$ in \Cref{eqn:guidance} for the general annotator model to guard the reliability of tool-augmented annotation. Specifically, before annotation, we train a set of tool-specific binary classifiers denoted as $ \mathcal{E} = \{\mathcal{E}_1, \mathcal{E}_2, \ldots, \mathcal{E}_K\} $, where each $ \mathcal{E}_k $ corresponds to a visual tool (excluding the Zoom-In Tool) and learns to predict the probability that a tool output contains spoof-related artifacts. Formally, for a tool result $ \mathcal{T}_k(\mathcal{I}) $, the expert prediction is computed as:
\begin{equation}
p_k = \mathcal{E}_k(\mathcal{T}_k(\mathcal{I})), \quad p_k \in [0, 1],
\end{equation}
where $ p_k $ represents the confidence that spoof traces exist in the given tool output. For each given tool-call request $ \mathcal{T}_k$, the tool-result $\mathcal{T}_k(\mathcal{I})$ will be fed into its expert network $ \mathcal{E}_k $. The resulting prediction $ p_k $ is then converted into a textual guidance $\mathcal{G}[\mathcal{T}_k(\mathcal{I})]$, such as ``\textit{This is the result of FFTTool. The expert predicts 87\% there's spoof trace}''. This mechanism allows the general annotator to combine expert-model-derived confidence with human interpretability, thereby producing more reliable tool-augmented reasoning.

\subsubsection{Data Verification and Re-Annotation}
To ensure the reliability of the annotated dataset, we verify data from three perspective (as shown in \Cref{fig:annotation}-(b)):
\begin{itemize}
    \item \textit{Correctness Verification.} Check whether the final reasoning answer matches the ground-truth label.
    \item \textit{Format Verification.} Ensure that tool calls follow the predefined API schema and analysis do not leak the ground-truth hint or expert confidence.
    \item \textit{Manual Verification.} Human experts validate the logical consistency between tool analyses and final conclusions.
\end{itemize}
Samples failing the above checks are re-annotated once, and those that still fail are marked as bad cases and discarded.

\subsection{Tool-Aware FAS Training Pipeline}

As illustrated in \Cref{fig:method}-(b), the overall training pipeline comprises three stages: FAS knowledge transfer, multi-turn tool-call format injection, and Diverse-Tool Group Relative Policy Optimization (DT-GRPO). The well trained TAR-FAS can conduct fine-grained visual reasoning via tool invocations, leading to interpretable and reliable final decisions, as shown in \Cref{fig:method}-(a).

\subsubsection{FAS Knowledge Transfer}

To endow the general MLLM with fundamental FAS knowledge, we leverage data annotated in the same format as I-FAS~\cite{ifas}, denoted as $\mathcal{D}_1 = \{ (\boldsymbol{q}^{(0)}, \boldsymbol{s}^{(0)})_i \}_{i=1}^{N_1}$ where $\boldsymbol{q}^{(0)}$ denotes the input image and user query, and $\boldsymbol{s}^{(0)}$ represents the target output sequence. The optimizing objective can be formulated as:
\begin{equation}
\mathcal{L}_{1} = - \mathbb{E}_{(\boldsymbol{q}, \boldsymbol{s}) \sim \mathcal{D}_{1}} \sum_{t=1}^{T^{(0)}} \log \pi_\theta(\boldsymbol{s}^{(0)}_t \mid \boldsymbol{q}^{(0)}, \boldsymbol{s}^{(0)}_{<t}),
\end{equation}
where $\pi_\theta$ denotes the token distribution from the current model, $T^{(0)}$ denotes the length of $\boldsymbol{s}^{(0)}$. Thus, we establish an initial alignment between visual and language in the FAS domain, resulting in an FAS expert MLLM.

\subsubsection{Tool-call Format Injection}

To equip the FAS expert MLLM with the tool-calling format, we train it on the ToolFAS-16K dataset, denoted as $\mathcal{D}_2 = \{ (\boldsymbol{q}^{(0)}, \boldsymbol{s}^{(0)}, \cdots, \boldsymbol{q}^{(L_i)}, \boldsymbol{s}^{(L_i)}) \}_{i=1}^{N_2}$. For each sample in $\mathcal{D}_2$, the first-round pair $(\boldsymbol{q}^{(0)}, \boldsymbol{s}^{(0)})$ is identical to that in $\mathcal{D}_1$ and serves as a prefix, enabling fast classification before the model proceeds to multi-turn, tool-based reasoning. We optimize the model by maximizing the generation likelihood with a loss scale on first-round generation:
\begin{equation}
\begin{aligned}
& \mathcal{L}_{nll}(l) = - \sum_{t=1}^{T_l} 
\log \pi_\theta(\boldsymbol{s}^{(l)}_{t}, | \boldsymbol{q}^{(<l)}, \boldsymbol{s}^{(<l)}, \boldsymbol{q}^{(l)}, \boldsymbol{s}^{(l)}_{<t}), \\
& \mathcal{L}_{2} = \mathbb{E}_{(\boldsymbol{q}, \boldsymbol{s}) \sim \mathcal{D}_2} \Big[ \alpha \cdot \mathcal{L}_{nll}(0) + (1 - \alpha) \cdot \sum_{l=1}^{L} \mathcal{L}_{nll}(l) ] \Big],
\end{aligned}
\end{equation}
where $L$ denotes the number of turns in each sampled data pair, $\alpha$ denotes a hyper-parameter which control the loss ratio of first-round generation. Applying a loss scale factor to the first-round generation prevents degradation of basic classification ability during long multi-turn training, enabling the model to retain its core classification skill while learning the tool-augmented reasoning format. Consequently, we inject the multi-turn tool-call format into the FAS expert MLLM, yielding a FAS Tool Learner MLLM.

\begin{table*}[t]
\small
    \centering
    \caption{Comparison in Protocol 2, illustrating the challenge of training solely on the CelebA-Spoof dataset followed by testing across 11 distinct datasets. We run each experiment 3 times under different seeds and report the average HTER and AUC.}
    \label{tab:p2}
    \begin{subtable}[t]{0.23\linewidth}
    \caption{\textbf{Average Over 11 Datasets}}
    \scalebox{0.94}{
    \begin{tabular}{p{10mm}<{\centering}p{10mm}<{\centering}p{10mm}<{\centering}}
    \toprule
    Methods & HTER(\%) & AUC \\
    \midrule
    ViTAF & 23.85 & 82.82  \\
    ViT-L & 21.08 & 85.61  \\
    FLIP & 18.73 & 87.90  \\
    I-FAS & 11.30 & 93.71 \\
    Ours & \textbf{7.54} & \textbf{96.67} \\
    \bottomrule
    \end{tabular}}
    \end{subtable}
    ~
    \centering
    \begin{subtable}[t]{0.23\linewidth}
    \caption{\textbf{CASIA-MFSD}}
    \scalebox{0.94}{
    \begin{tabular}{p{10mm}<{\centering}p{10mm}<{\centering}p{10mm}<{\centering}}
    \toprule
    Methods & HTER(\%) & AUC \\
    \midrule
    ViTAF & 3.11 & 99.48  \\
    ViT-L & 0.93 & 99.95  \\
    FLIP & 4.88 & 98.48  \\
    I-FAS & 1.11 & 99.88 \\
    Ours & \textbf{0.00} & \textbf{100.00} \\
    \bottomrule
    \end{tabular}}
    \end{subtable}
    ~
    \centering
    \begin{subtable}[t]{0.23\linewidth}
    \caption{\textbf{CASIA-SURF-3DMask}}
    \scalebox{0.94}{
    \begin{tabular}{p{10mm}<{\centering}p{10mm}<{\centering}p{10mm}<{\centering}}
    \toprule
    Methods & HTER(\%) & AUC \\
    \midrule
    ViTAF & 32.44 & 75.20  \\
    ViT-L & 23.54 & 84.22  \\
    FLIP & 8.83 & 96.93  \\
    I-FAS & 6.18 & 98.40 \\
    Ours & \textbf{2.09} & \textbf{99.65} \\
    \bottomrule
    \end{tabular}}
    \end{subtable}
    ~
    \centering
    \begin{subtable}[t]{0.23\linewidth}
    \caption{\textbf{HKBU-MARs-V1+}}
    \scalebox{0.94}{
    \begin{tabular}{p{10mm}<{\centering}p{10mm}<{\centering}p{10mm}<{\centering}}
    \toprule
    Methods & HTER(\%) & AUC \\
    \midrule
    ViTAF & 49.29 & 57.28  \\
    ViT-L & 33.33 & 73.88  \\
    FLIP & 17.25 & 88.31  \\
    I-FAS & 18.64 & 88.77 \\
    Ours & \textbf{3.48} & \textbf{99.71} \\
    \bottomrule
    \end{tabular}}
    \end{subtable}
    ~
    \centering
    \begin{subtable}[t]{0.23\linewidth}
    \caption{\textbf{HiFiMask}}
    \scalebox{0.94}{
    \begin{tabular}{p{10mm}<{\centering}p{10mm}<{\centering}p{10mm}<{\centering}}
    \toprule
    Methods & HTER(\%) & AUC \\
    \midrule
    ViTAF & 37.30 & 67.10  \\
    ViT-L & 32.81 & 72.58  \\
    FLIP & 28.32 & 76.50  \\
    I-FAS & 28.23 & 77.17 \\
    Ours & \textbf{17.97} & \textbf{90.23} \\
    \bottomrule
    \end{tabular}}
    \end{subtable}
    ~
    \centering
    \begin{subtable}[t]{0.23\linewidth}
    \caption{\textbf{MSU-MFSD}}
    \scalebox{0.94}{
    \begin{tabular}{p{10mm}<{\centering}p{10mm}<{\centering}p{10mm}<{\centering}}
    \toprule
    Methods & HTER(\%) & AUC \\
    \midrule
    ViTAF & 12.86 & 93.14  \\
    ViT-L & 20.87 & 85.65  \\
    FLIP & 19.37 & 89.95  \\
    I-FAS & \textbf{5.63} & \textbf{98.73} \\
    Ours & 5.71 & 98.29 \\
    \bottomrule
    \end{tabular}}
    \end{subtable}
    ~
    \centering
    \begin{subtable}[t]{0.23\linewidth}
    \caption{\textbf{OULU-NPU}}
    \scalebox{0.94}{
    \begin{tabular}{p{10mm}<{\centering}p{10mm}<{\centering}p{10mm}<{\centering}}
    \toprule
    Methods & HTER(\%) & AUC \\
    \midrule
    ViTAF & 26.73 & 81.28  \\
    ViT-L & 29.42 & 78.07  \\
    FLIP & 20.57 & 87.30  \\
    I-FAS & 14.86 & \textbf{92.68} \\
    Ours & \textbf{14.45} & 92.63 \\
    \bottomrule
    \end{tabular}}
    \end{subtable}
    ~
    \centering
    \begin{subtable}[t]{0.23\linewidth}
    \caption{\textbf{REPLAY-ATTACK}}
    \scalebox{0.94}{
    \begin{tabular}{p{10mm}<{\centering}p{10mm}<{\centering}p{10mm}<{\centering}}
    \toprule
    Methods & HTER(\%) & AUC \\
    \midrule
    ViTAF & 12.38 & 95.73  \\
    ViT-L & 16.58 & 92.00  \\
    FLIP & 25.67 & 81.37  \\
    I-FAS & 9.15 & 95.12 \\
    Ours & \textbf{5.75} & \textbf{96.69} \\
    \bottomrule
    \end{tabular}}
    \end{subtable}
    ~
    \centering
    \begin{subtable}[t]{0.23\linewidth}
    \caption{\textbf{Rose-Youtu}}
    \scalebox{0.94}{
    \begin{tabular}{p{10mm}<{\centering}p{10mm}<{\centering}p{10mm}<{\centering}}
    \toprule
    Methods & HTER(\%) & AUC \\
    \midrule
    ViTAF & 69.34 & 74.22  \\
    ViT-L & 80.47 & 71.69  \\
    FLIP & 80.73 & 73.60  \\
    I-FAS & 5.52 & 98.48 \\
    Ours & \textbf{3.61} & \textbf{99.16} \\
    \bottomrule
    \end{tabular}}
    \end{subtable}
    ~
    \centering
    \begin{subtable}[t]{0.23\linewidth}
    \caption{\textbf{SIW}}
    \scalebox{0.94}{
    \begin{tabular}{p{10mm}<{\centering}p{10mm}<{\centering}p{10mm}<{\centering}}
    \toprule
    Methods & HTER(\%) & AUC \\
    \midrule
    ViTAF & 14.74 & 92.51  \\
    ViT-L & 9.03 & 96.56  \\
    FLIP & 11.01 & 95.40  \\
    I-FAS & \textbf{4.02} & \textbf{98.34} \\
    Ours & 8.35 & 96.93 \\
    \bottomrule
    \end{tabular}}
    \end{subtable}
    ~
    \centering
    \begin{subtable}[t]{0.23\linewidth}
    \caption{\textbf{SIW-M-V2}}
    \scalebox{0.94}{
    \begin{tabular}{p{10mm}<{\centering}p{10mm}<{\centering}p{10mm}<{\centering}}
    \toprule
    Methods & HTER(\%) & AUC \\
    \midrule
    ViTAF & 26.72 & 80.70  \\
    ViT-L & 17.26 & 90.37  \\
    FLIP & 25.95 & 80.78  \\
    I-FAS & \textbf{10.89} & \textbf{95.02} \\
    Ours & 11.72 & 94.75 \\
    \bottomrule
    \end{tabular}}
    \end{subtable}
    ~
    \centering
    \begin{subtable}[t]{0.23\linewidth}
    \caption{\textbf{WMCA}}
    \scalebox{0.94}{
    \begin{tabular}{p{10mm}<{\centering}p{10mm}<{\centering}p{10mm}<{\centering}}
    \toprule
    Methods & HTER(\%) & AUC \\
    \midrule
    ViTAF & 29.88 & 77.14  \\
    ViT-L & 34.39 & 75.13  \\
    FLIP & 19.36 & 88.73  \\
    I-FAS & 20.07 & 89.17 \\
    Ours & \textbf{9.78} & \textbf{95.29} \\
    \bottomrule
    \end{tabular}}
    \end{subtable}
\end{table*}

\subsubsection{Diverse-Tool GRPO}

To enable the model to autonomously learn efficient tool-use, we introduce DT-GRPO. We only use query-label pairs from CelebA-Spoof dataset denoted as $\mathcal{D}_3 = \{ (\boldsymbol{q}, \boldsymbol{a})_i \}_{i=1}^{N_3}$ where $\boldsymbol{a}$ denotes the binary label. For each given query $\boldsymbol{q}$, DT-GRPO samples $G$ responses $\{ o_1, o_2, \cdots, o_G \}$ using the current policy model $\pi_{\theta_{old}}$. The reward of each response $\{ R_{1}, R_{2}, \cdots, R_{G} \}$ is calculated through tool-diversity reward function. The on-policy training objective can be formulated as:
\begin{equation}
\begin{aligned}
\mathcal{L}_3 = & - \mathbb{E}_{(\boldsymbol{q}, \boldsymbol{s}) \sim \mathcal{D}_3, \{ o_i \}_{i=1}^G \sim \pi_{\theta_{old}} (\cdot | \boldsymbol{q})} \\
& \bigg[ \frac{1}{G} \sum_{i=1}^G \frac{1}{|o_i|} \sum_{t=1}^{|o_i|} \bigg( \frac{\pi_\theta (o_{i,t} | q, o_{i, <t})}{\pi_{\theta_{old}} (o_{i,t} | q, o_{i, <t})} \cdot A_{i, t} \bigg) \bigg],
\end{aligned}
\end{equation}
where
\begin{equation}
A_{i, t} = \frac{R_i - mean(\{ R_1, \cdots, R_G \})}{std(\{ R_1, \cdots, R_G \})}. \notag
\end{equation}

For each rollout response $o_i$, the tool-diversity reward $R_i$ for is evaluated from three perspective:

\noindent \textbf{Fast answer reward.} A fast classification will be given in the first-turn and the total fast reward can be formulated as: 
\begin{equation}
    R_{fast} = R_{fast}^{fmt} + \mathbb{I}_{R_{fast}^{fmt} > -1} \cdot \mathbb{I}_{CLS_{fast}=label} \, \, \, \, ,
\end{equation}
where $R_{fast}^{fmt}$ is $-1$ if the format is wrong otherwise $0$, and $CLS_{fast}$ denotes the fast classification.

\noindent \textbf{Reasoning reward.} The reasoning reward constrain the final accuracy and reasoning format formulated as:
\begin{equation}
    R_{rsn} = R_{rsn}^{fmt} + \mathbb{I}_{R_{rsn}^{fmt} > -1} \cdot \mathbb{I}_{CLS_{final}=label} \, \, \, \, ,
\end{equation}
where $R_{rsn}^{fmt}$ is $-1$ if a wrong reasoning format or a invalid tool-call is presented otherwise $0$, and $CLS_{final}$ denotes the final decision.

\noindent \textbf{Tool reward.} 
We aim to encourage the model to utilize various tools in order to achieve a correct classification result. For a rollout that reaches a final decision in $L$ turns, the model will select one of the $K$ available tools during turns $2$ to $(L-1)$, and produces the final answer at turn $L$. Let $v^{(l)} \in \{0,1\}$ indicate a valid tool-call and $\gamma_k$ indicate tool weight for tool $\mathcal{T}_k$, the tool-call diversity score is defined as:
\begin{equation}
F_{tool} = \sum_{k=1}^{K} \gamma_k  \cdot \max \bigg[ \sum_{l=2}^{L} v^{(l)} \, \mathbb{I}_{\mathcal{T}^{(l)} = \mathcal{T}_k} , \,\, 1 \bigg] .
\end{equation}
We gate this reward by the final outcome:
\begin{equation}
R_{\text{tool}}  =F_{tool} \cdot 
\mathbb{I}_{CLS_{final}=label}.
\end{equation}

\noindent \textbf{Total reward.}
The total reward is formulated as:
\begin{equation}
R = \beta_{fast} \cdot R_{fast} + \beta_{rsn} \cdot R_{rsn} + \beta_{tool} \cdot R_{tool}.
\end{equation}

\section{Experiments}

\subsection{Experimental Setup}

\noindent \textbf{Databases and Protocols.} To evaluate the generalization and robustness of our method, we choose the most challenging cross-domain protocol following I-FAS \cite{ifas}. We train on a single source domain CelebA-Spoof \cite{celebaspoof} and perform cross-domain testing on 11 target domains including MSU-MFSD \cite{msumfsd}, CASIA-MFSD \cite{casiafasd}, Idiap Replay Attack \cite{replayattack}, OULU-NPU \cite{oulunpu}, SIW \cite{siw}, Rose-Youtu \cite{ksa}, HKBU-MARs-V1+ \cite{hkbu}, WMCA \cite{wmca}, SIW-M-V2 \cite{siwmv2}, CASIA-SURF-3DMask \cite{nasfas} and HiFiMask \cite{hifimask}. The result of traditional ICMO protocol is shown in supplementary materials.

\noindent \textbf{Evaluation Metrics.}
In line with the evaluation principles, we use HTER and AUC to assess the model’s performance. (1) HTER is a measure of the false rejection and false acceptance error rates, and its value is taken as the average of the false rejection rate (FRR) and the false acceptance rate (FAR). (2) AUC measures the algorithm’s overall performance, and its value represents the area under the ROC curve. Notably, we preserve the original logits of the classification token in final answer throughout the MLLM generation process for metric computation.

\noindent \textbf{Implementation Details.} We align, crop, and resize face images to $224 \times 224$. We use InternVL-3-8B \cite{internvl3} as the base MLLM model. For knowledge transfer and format injection, we use LoRA (rank=16) finetuning with learning rate 1e-4 and loss scale factor $\alpha$=0.9, using a batchsize of 256 for up to 4000 steps. For DT-GRPO, we use full-parameter finetuning with learning rate 1e-6, KL coefficient 0.0. We set the reward coefficient $\beta_{fast}$=0.1, $\beta_{rsn}$=0.5, $\beta_{tool}$=0.4, $\gamma_{k}$=0.2. We configure the training process with a completion batchsize of 512 (rollout number G=8) and a maximum of 300 steps. AdamW with weight decay 0.05 is used throughout.
Note that although InternVL preprocesses inputs to $448 \times 448$, our input resolution remains $224 \times 224$.

\subsection{Comparison Results}
To evaluate the generalizability of our approach, we choose the most challenging one-to-eleven protocol \cite{ifas} utilizing a single dataset (CelebA-Spoof) as the source domain and perform cross-domain testing across 11 distinct datasets. We compare our method with ViT-L \cite{vit}, ViTAF \cite{vitfas}, FLIP \cite{flip} and I-FAS \cite{ifas}. As shown in \Cref{tab:p2}, our method demonstrates a compelling advantage of near 3\% HTER to the previous SOTA techniques. This scenario reflects real-world challenges that require robustness against emerging attacks and shifts. Significantly, the datasets CASIA-SURF-3DMask and HKBU-MARs-V1+ encompass attack modalities absent in the source domain, including sophisticated 3D attacks and novel material-based attacks. Under these challenging condition, our method significantly outperforms previous FAS approaches by a substantial margin, demonstrating the superior generalizability gained through tool-augmented reasoning.

\subsection{Ablation Study}

\begin{table}[h]
\small
\centering
\caption{Ablation study of different visual tools. Frequency tools denotes FFT and Wavelet, texture tools denotes LBP, structure tools denotes HOG and Edge Detection.}
\label{tab:tools}
\resizebox{0.9\columnwidth}{!}{
\begin{tabular}{p{40mm}<{}p{15mm}<{\centering}p{15mm}<{\centering}}
\toprule
\multicolumn{1}{c}{\multirow{2}{*}{Tools}}                            & \multicolumn{2}{c}{Results}  \\ \cmidrule(lr){2-3} 
\multicolumn{1}{c}{}                                                  & \multicolumn{1}{c}{HTER(\%)} & \multicolumn{1}{c}{AUC} \\ \midrule
ZoomIn Tool  & 10.63 & 93.90 \\
ZoomIn + Frequency Tools  & 9.55 & 95.40 \\
ZoomIn + Texture Tools & 9.90 & 95.75 \\
ZoomIn + Structure Tools & 8.71 & 96.32 \\
\rowcolor[HTML]{C0C0C0} All Tools & \textbf{7.54} & \textbf{96.67} \\ [-0.6ex] \bottomrule
\end{tabular}
}
\end{table}

\noindent \textbf{Effectiveness of different visual tools.} To demonstrate the generalization improvement brought by visual tools, we conducted extensive comparative experiments, as shown in the \Cref{tab:tools}. Available tools (except ZoomIn) are categorized into three major types: (1) Frequency tools, which extract frequency-domain information from images, including FFT and Wavelet; (2) Texture tools, which analyze facial texture patterns, represented by LBP; and (3) Structure tools, which capture structural information within images, including HOG and EdgeDetection. The model that only employs ZoomIn Tool serves as the baseline. When Frequency, Texture, or Structure tools are individually incorporated, each configuration yields performance gains over the baseline. Furthermore, when all tools are available for the model, the overall performance is further enhanced, confirming the necessity of these visual tools.

\begin{table}[t]
\small
\centering
\caption{Ablation study of different training strategy. `KT' denotes FAS Knowledge Transfer, `FI' denotes tool-call Format Injection, and 'RL' denotes DT-GRPO.}
\label{tab:train}
\resizebox{0.9\columnwidth}{!}{
\begin{tabular}{p{10mm}<{\centering}p{10mm}<{\centering}p{10mm}<{\centering}p{15mm}<{\centering}p{15mm}<{\centering}}
\toprule
\multicolumn{3}{c}{Training Stage}  & \multicolumn{2}{c}{Results}                            \\ \cmidrule(lr){1-3} \cmidrule(lr){4-5}
\multicolumn{1}{c}{KT} & FI & RL  & \multicolumn{1}{c}{HTER(\%)} & \multicolumn{1}{c}{AUC} \\ \midrule
            &           & \ding{51} & 20.79 & 86.43 \\
            & \ding{51} &           & 16.80 & 89.65 \\ 
 \ding{51}  & \ding{51} &           & 13.29 & 91.46 \\
            & \ding{51} & \ding{51} & 9.23  & 95.11 \\
\rowcolor[HTML]{C0C0C0} \ding{51}  & \ding{51} & \ding{51} & \textbf{7.54} & \textbf{96.67} \\ [-0.6ex] \bottomrule
\end{tabular}}
\end{table}

\begin{figure*}[t]
    \centering
    \includegraphics[width=0.96\linewidth]{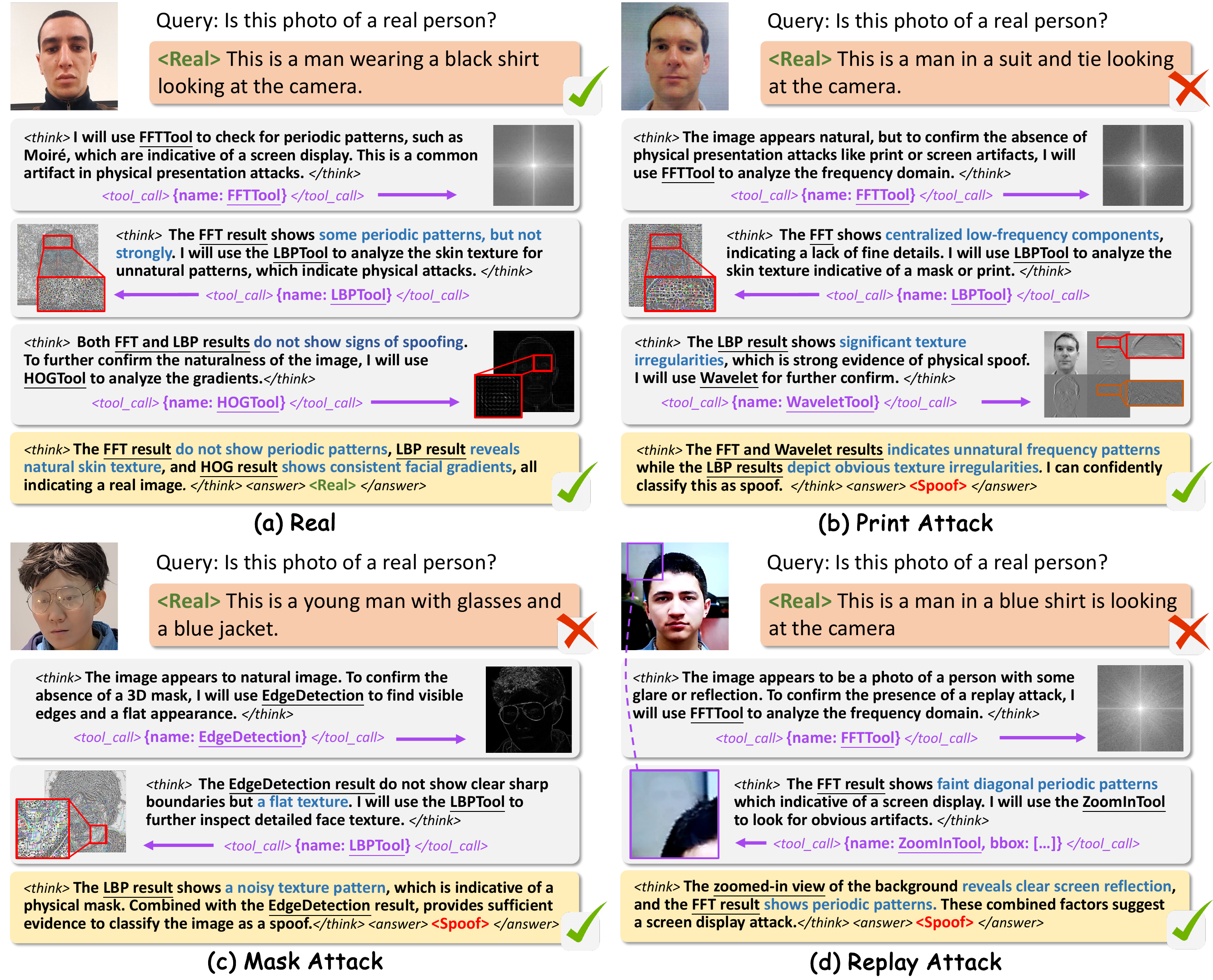}
    \caption{The visualization of TAR-FAS results. We select one sample for each type: real, print, replay, and mask attacks. The zoom in operation in red boxes is for better understandings.}
    \label{fig:vis}
\vspace{-5pt}
\end{figure*}

\noindent \textbf{Effectiveness of each training stages.} To validate the effectiveness of the proposed training paradigm and the ToolFAS-16K dataset, we conducted comprehensive experiments exploring all possible combinations that the model can perform reasoning-based classification. The results are shown in the \Cref{tab:train} where `KT' denotes FAS Knowledge Transfer, `FI' denotes tool-call Format Injection, and 'RL' denotes DT-GRPO. The results lead to the following conclusions: (1) The proposed DT-GRPO enables the model to autonomously learn efficient tool-use, significantly enhancing generalization on FAS classification tasks. FI+RL improves over FI by 5.46\% HTER, and KT+FI+RL surpasses FI+RL by 5.21\% HTER as well. (2) The FI process based on the ToolFAS-16K dataset is indispensable. Without FI step, the RL stage fails to train the model effectively, resulting in an HTER of only 20.79\%. (3) The KT process can effectively transfer FAS domain knowledge into the pretrained general MLLM. KT+FI outperforms FI by 1.81\% HTER, while KT+FI+RL further improves upon FI+RL by another 1.56\% HTER. In summary, these results confirm the critical role of the proposed tool-aware FAS training paradigm and the ToolFAS-16K dataset.

\begin{table}[h]
\small
\centering
\caption{Ablation study of different resolution and backbone.}
\label{tab:backbone}
\resizebox{0.9\columnwidth}{!}{
\begin{tabular}{p{25mm}<{}p{12mm}<{\centering}p{15mm}<{\centering}p{15mm}<{\centering}}
\toprule
\multicolumn{1}{c}{\multirow{2}{*}{Backbone}} & \multicolumn{1}{c}{\multirow{2}{*}{Resize}} & \multicolumn{2}{c}{Results} \\ \cmidrule(lr){3-4} 
\multicolumn{1}{c}{} & & \multicolumn{1}{c}{HTER(\%)} & \multicolumn{1}{c}{AUC} \\ \midrule
Qwen2.5-VL-7B     &           & 9.45  & 95.32 \\
Qwen2.5-VL-7B     & \ding{51} & 8.60  & 96.08 \\
\rowcolor[HTML]{C0C0C0} InternVL3-8B    & \ding{51} & \textbf{7.54} & \textbf{96.67} \\ [-0.6ex] \bottomrule
\end{tabular}}
\end{table}

\noindent \textbf{Influence of backbone and image resolution.} Since the InternVL model preprocesses images into multiple $448 \times 448$ patches during pretraining, our $224 \times 224$ input images are also resized to $448 \times 448$, leading to an unfair comparison. To address this issue, we conducted additional experiments on the Qwen2.5-VL \cite{qwen25vl} with two settings, direct $224$ input and $224$ resized $448$ input, to verify the effectiveness of TAR-FAS, as shown in \Cref{tab:backbone}. The results indicate that resizing $224 \times 224$ images to $448 \times 448$ provides a slight performance improvement; however, this gain is marginal compared with the enhancement achieved by our proposed method. Moreover, the results demonstrate that TAR-FAS achieves state-of-the-art (SOTA) performance on both the InternVL and QwenVL series models, further validating the effectiveness and robustness of our approach.

\subsection{Visualization}
When different spoof types of samples are presented to TAR-FAS, the model adaptively invokes different tools based on its initial observation. We visualize one example for each spoof type (real, print, replay, and mask attacks) in \Cref{fig:vis}. As shown, TAR-FAS prefers frequency tools for print and replay attacks, structure tools for mask attacks, while texture tools provide benefits across all attack types. Moreover, TAR-FAS can reconsider and overturn its initial incorrect predictions based on the tool-augmented investigation, leading to more accurate final decisions. These results demonstrate that integrating visual tools guide MLLM to perceive subtle spoof cues, thus enhances the generalization and interpretability.
\section{Conclusion}

In this work, we proposed TAR-FAS, a novel framework that reformulates the FAS task as a CoT-VT paradigm by integrating visual tool into MLLMs. To support tool-augmented learning, we constructed ToolFAS-16K, a large-scale dataset containing multi-turn tool-use trajectories generated through a dedicated annotation pipeline. We further introduced a tool-aware FAS training strategy including a DT-GRPO which enable the model to automatically learn efficient tool-use. Extensive experiments verify the effectiveness of TAR-FAS, achieving SOTA performance across multiple benchmarks. We believe that continued optimization and the development of more powerful visual tools will further advance FAS toward higher levels of generalization and interpretability.
{
    \small
    \bibliographystyle{ieeenat_fullname}
    \bibliography{main}
}

\appendix
\clearpage
\setcounter{page}{1}
\maketitlesupplementary
\renewcommand{\thefigure}{S\arabic{figure}}
\renewcommand{\thetable}{S\arabic{table}}
\renewcommand{\theequation}{S\arabic{equation}}

\section{Data Annotation Pipeline}

\subsection{Data Selection}

\begin{figure}[t]
    \centering
    \includegraphics[width=\linewidth]{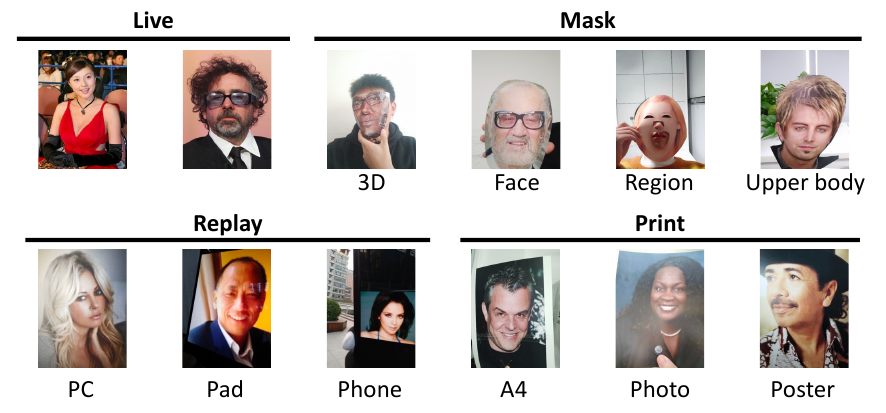}
    \caption{The detailed spoof types in ToolFAS-16K.}
    \label{fig:supp_data_description}
\end{figure}

In the ToolFAS-16K dataset, we cover the ten spoof types in CelebA-Spoof \cite{celebaspoof}. The sample of each spoof type is shown in \Cref{fig:supp_data_description}.

\subsection{Annotation System Prompt Design}

To ensure consistent reasoning behavior during annotation, we design a structured system prompt $ \mathcal{S} $ that explicitly defines the model’s role, objectives, and interaction rules. Formally, the system prompt is composed of five components:
\begin{equation*}
\mathcal{S} = \{ \mathcal{P}_{\text{role}}, \mathcal{P}_{\text{principle}}, \mathcal{P}_{\text{workflow}}, \mathcal{P}_{\text{tools}}, \mathcal{P}_{\text{conclusion}} \},
\end{equation*}
where $ \mathcal{P}_{\text{role}} $ specifies the model’s \textit{Role \& Mission}, $ \mathcal{P}_{\text{principle}} $ defines the \textit{Core Principles} for reasoning, $ \mathcal{P}_{\text{workflow}} $ encodes the \textit{Behavioral Rules} for multi-turn interactions, $ \mathcal{P}_{\text{tools}} $ enumerates the \textit{Available Tools}, and $ \mathcal{P}_{\text{conclusion}} $ provides instructions for generating the \textit{Final Conclusion}. This structured design ensures that the annotator model follows a stable reasoning pattern alternating between internal thought and tool invocation. The complete system prompt is shown in \Cref{fig:system}.

\subsection{Expert Model}

In our experiments, these expert models achieve around 90\% accuracy on the CelebA-Spoof domain without using RGB inputs, and each shows distinctive sensitivity to certain spoof types, validating their reliability.

\begin{figure*}[t]
\centering
\includegraphics[width=\linewidth]{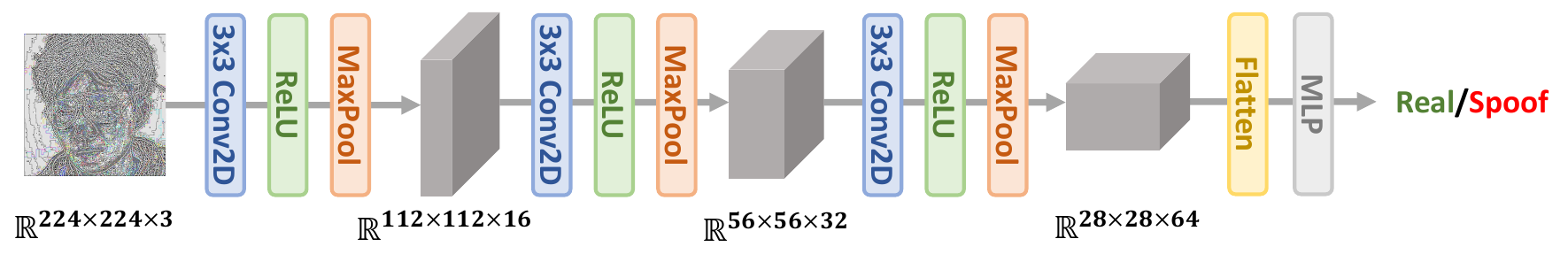}
\caption{The detailed architecture of expert model for each tool. The input of expert model is only tool result (without RGB image). The `MaxPool' perform double downsampling. The `MLP' denotes two linear layer with a ReLU achitation function.}
\label{fig:expert_model}
\end{figure*}

\noindent \textbf{Architecture.} We use a simple CNN architecture to train expert model $\{ \mathcal{E}_1, \cdots , \mathcal{E}_K \}$ for each tool with only the tool result as input and the binary classification as output. The detailed architecture is illustrated in \Cref{fig:expert_model}.

\noindent \textbf{Implementation.} We use the CelebA-Spoof dataset~\cite{celebaspoof} to train the expert model for each tool. All images are aligned, cropped, and resized to $224 \times 224$. We randomly select 5,000 identities from CelebA-Spoof and split them into 4,000 for training and 1,000 for testing. All samples belonging to each identity are included. The models are trained for 10 epochs using the Adam optimizer with a learning rate of 0.001.

\begin{figure}[t]
    \centering
    \includegraphics[width=\linewidth]{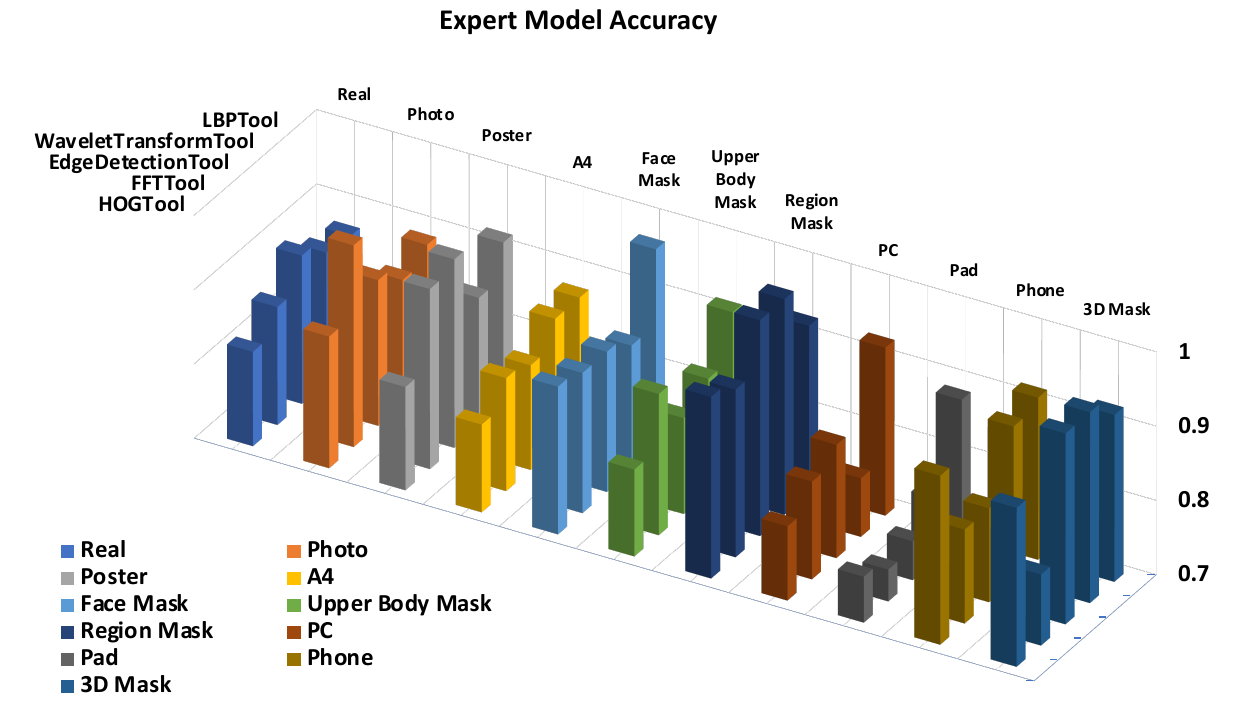}
    \caption{The detailed accuracy of expert model for each tool.}
    \label{fig:expert_model_acc}
\end{figure}

\noindent \textbf{Accuracy.} We evaluate each expert model using a fixed threshold of 0.5, and the results are presented in \Cref{fig:expert_model_acc}. As shown, the average accuracy of all tools exceeds 80\%, demonstrating that even when only tool results are used as input, they possess considerable discriminative and generalization capabilities. Moreover, different visual tools show varying effectiveness across spoof types, highlighting the necessity of employing diverse visual tools.

\begin{figure*}
\begin{tcolorbox}[
    colback=white,          
    colframe=black,         
    title=\textbf{System Prompt}, 
    boxrule=0.8pt,          
    arc=2mm,                
    fonttitle=\normalsize\bfseries,
    colbacktitle=black!5,   
    coltitle=black,         
    left=3mm, right=3mm, top=2mm, bottom=2mm
]
\#\# Role \& Mission \\
You are a face forensics expert. Your mission is to classify an image as either 'real' or 'spoof' by analyzing evidence strictly within the **facial region**, focusing **only on physical presentation attacks** (e.g., printed photos, screen displays). \\
You will be given a hint early in the conversation. Do not mention the hint when making your decision. Your final classification must match the hint, and be supported by image evidence and tool-based analysis. \\

\#\# Core Principle \\
Concentrate solely on **physical attack artifacts**. These include:
- **Semantic clues**: unnatural flatness, rigid expressions, lack of 3D structure, unnatural reflections. \\
- **Pixel-level clues**: print texture, Moiré patterns, screen glare, paper/screen surface noise. \\
Since input images are cropped and aligned, **do not consider black borders and compression artifacts as spoof clues**. \\

\#\# Workflow \& Behavioral Rules \\
1. Begin with a brief visual scan of the **facial region**. **Ignore black borders and all context outside the face.** \\
2. If needed, call **ONE tool at a time** to test a specific hypothesis, either to look for signs of physical attack, or to confirm their absence. \\
- Each tool request must include a clear expectation (what you're testing for).
3. When you receive tool results, you may receive an **Expert Judgment** on the result. \\
- You may consider the expert's interpretation as a reference, but **must perform your own independent analysis**. \\
- Your reasoning should not blindly follow the expert; only adopt it when it aligns with your observations. \\
4. When you are confident, provide your conclusion. \\

\#\# Available Tools \\
- **ZoomInTool**: Inspect local details for physical (print/screen) or digital (blending) artifacts. \\
- **FFTTool (Fast Fourier Transform):** Visualizes the image's frequency domain. Used to detect periodic patterns like screen Moiré effects or subtle artifacts from digital generation. \\
- **EdgeDetectionTool**: Find inconsistent edges from cutouts or digital blending. \\
- **LBPTool**: Analyze skin texture for unnatural or synthetic patterns. \\
- **WaveletTransformTool**: Find subtle digital tampering or noise mismatches via multi-scale analysis. \\
- **HOGTool**: Check facial structure gradients, which are often disrupted in physical attacks. \\

\#\# Final Conclusion \\
Your conclusion must be `Real' or `Spoof'.
\end{tcolorbox}
\caption{Complete System prompt.}
\label{fig:system}
\end{figure*}

\begin{table*}[t]
\small
\centering
\caption{Comparison with the closest and SOTA FAS methods in leave-one-out protocol on MSU-MFSD (M), CASIA-FASD (C), ReplayAttack (I), and OULU-NPU (O) datasets. Avg. indicates the average performance across four experimental scenarios. The scores presented in bold represent the best performance.}
\label{tab:p2}
\begin{tabular}{p{30mm}<{\centering}p{10mm}<{\centering}p{10mm}<{\centering}p{10mm}<{\centering}p{10mm}<{\centering}p{10mm}<{\centering}p{10mm}<{\centering}p{10mm}<{\centering}p{10mm}<{\centering}p{15mm}<{\centering}}
\toprule
\multirow{2}{*}{Methods} & \multicolumn{2}{c}{O\&C\&I to M}   & \multicolumn{2}{c}{O\&M\&I to C} & \multicolumn{2}{c}{O\&C\&M to I} & \multicolumn{2}{c}{I\&C\&M to O} & Avg. \\ 
\cmidrule(r){2-3}
\cmidrule(r){4-5}
\cmidrule(r){6-7}
\cmidrule(r){8-9}
\cmidrule{10-10}
 & HTER(\%) & AUC & HTER(\%) & AUC & HTER(\%) & AUC & HTER(\%) & AUC & HTER(\%) \\ 
\midrule
FGHV \cite{FGHV} & 9.17 & 96.92 & 12.47 & 93.47 & 16.29 & 90.11 & 13.58 & 93.55 & 12.88 \\
GDA \cite{GDA} & 9.20 & 98.00 & 12.20 & 93.00 & 10.00 & 96.00 & 14.40 & 92.60 & 11.45 \\
PatchNet \cite{patchnet} & 7.10 & 98.46 & 11.33 & 94.58 & 13.40 & 95.67 & 11.82 & 95.07 & 10.91 \\
SSAN \cite{ssanr} & 6.67 & 98.75 & 10.00 & 96.67 & 8.88 & 96.79 & 13.72 & 93.63 & 9.82 \\
IADG \cite{iadg} & 5.41 & 98.19 & 8.70 & 96.40 & 10.62 & 94.50 & 8.86 & 97.14 & 8.40 \\
UDG-FAS \cite{udgfas} & 5.95 & 98.47 & 9.82 & 96.76 & 5.86 & 98.62 & 10.97 & 95.36 & 8.15 \\
TTDG \cite{ttdg} & 4.16 & 98.48 & 7.59 & 98.18 & 9.62 & 98.18 & 10.00 & 96.15 & 7.84 \\
SA-FAS \cite{safas} & 5.95 & 96.55 & 8.78 & 95.37 & 6.58 & 97.54 & 10.00 & 96.23 & 7.83 \\
DiVT-M \cite{divtm} & 2.86 & 99.14 & 8.67 & 96.92 & 3.71 & 99.29 & 13.06 & 94.04 & 7.08 \\
GAC-FAS \cite{gacfas} & 5.00 & 97.56 & 8.20 & 95.16 & 4.29 & 98.87 & 8.60 & 97.16 & 6.52 \\
\midrule
FLIP \cite{flip} & 4.95 & 98.11 & 0.54 & 99.98 & 4.25 & 99.07 & 2.31 & 99.63 & 3.01 \\
CFPL-FAS \cite{cfpl} & 1.43 & 99.28 & 2.56 & 99.10 & 5.43 & 98.41 & 2.50 & 99.42 & 2.98 \\
I-FAS \cite{ifas} & \textbf{0.32} & \textbf{99.88} & 0.04 & 99.99 & 3.22 & 98.48 & \textbf{1.74} & 99.66 & 1.33 \\ 
\rowcolor[HTML]{C0C0C0} TAR-FAS (Ours) & 2.86 & 99.35 & \textbf{0.00} & \textbf{100.00} & \textbf{0.43} & \textbf{99.86} & 1.91 & \textbf{99.72} & \textbf{1.30} \\ [-0.6ex]
\bottomrule
\end{tabular}
\end{table*}

\section{Tool-Aware FAS Training Pipeline}

\subsection{Training Prompt Design}

\noindent \textbf{System Prompt.} The system prompt contains \textit{Role \& Mission} description, tool description in XML format. The \textit{Role \& Mission} description is shown as follows:
\begin{tcolorbox}[
    colback=white,          
    colframe=black,         
    title=\textbf{System Prompt}, 
    boxrule=0.8pt,          
    arc=2mm,                
    fonttitle=\normalsize\bfseries,
    fontupper=\small,
    colbacktitle=black!5,   
    coltitle=black,         
    left=3mm, right=3mm, top=2mm, bottom=2mm
]
\#\# Role \& Mission \\
You are a face forensics expert. Your mission is to classify an image as either `Real' or `Spoof' by analyzing evidence strictly within the **facial region**, focusing only on physical presentation attack.
\end{tcolorbox}
While, the tool description follows Json Schema and adopt the Hermes template used in Qwen2.5-VL \cite{qwen25vl}.

\noindent \textbf{First-Round Query.} The first-round query ask the model to classify whether the given image is of a real person.
\begin{tcolorbox}[
    colback=white,          
    colframe=black,         
    title=\textbf{First-Round Query}, 
    boxrule=0.8pt,          
    arc=2mm,                
    fonttitle=\normalsize\bfseries,
    colbacktitle=black!5,   
    coltitle=black,         
    left=3mm, right=3mm, top=2mm, bottom=2mm
]
Is this photo of a real person? (Do not use any tools)
\end{tcolorbox}

\noindent \textbf{Tool Guidance and Format Declaration.} In the second round, user prompt conatains tool guidance and format declaration.
\begin{tcolorbox}[
    colback=white,          
    colframe=black,         
    title=\textbf{Tool Guidance}, 
    boxrule=0.8pt,          
    arc=2mm,                
    fonttitle=\normalsize\bfseries,
    colbacktitle=black!5,   
    coltitle=black,         
    left=3mm, right=3mm, top=2mm, bottom=2mm
]
Wait, you should re-examine the image and give the final answer (use tools if needed).
\end{tcolorbox}
\begin{tcolorbox}[
    colback=white,          
    colframe=black,         
    title=\textbf{Format Declaration}, 
    boxrule=0.8pt,          
    arc=2mm,                
    fonttitle=\normalsize\bfseries,
    fontupper=\small,
    colbacktitle=black!5,   
    coltitle=black,         
    left=3mm, right=3mm, top=2mm, bottom=2mm
]
Think first, call tools if needed, then answer. Format strictly as:  \texttt{<}think\texttt{>} ... \texttt{<}/think\texttt{>} \texttt{<}tool\_call\texttt{>} ... \texttt{<}/tool\_call\texttt{>} (if tools needed) \texttt{<}answer\texttt{>}(\texttt{<}Spoof\texttt{>}/\texttt{<}Real\texttt{>})\texttt{<}/answer\texttt{>}
\end{tcolorbox}
After the second round, the user prompt will contain a tool result image and format declaration.

\subsection{DT-GRPO Format Constrain}


\noindent \textbf{Fast answer format.} A classification will be give in the first-turn with a strict format of \texttt{<}CLS\texttt{>}\texttt{<}reason\texttt{>}\texttt{<}/reason\texttt{>}. A wrong format will cause to a penalty of $-1$.

\noindent \textbf{Reasoning format.} The reasoning format for each turn should be \texttt{<}think\texttt{>}\texttt{<}/think\texttt{>} + \texttt{<}tool\_call\texttt{>}\texttt{<}/tool\_call\texttt{>} or \texttt{<}think\texttt{>}\texttt{<}/think\texttt{>} + \texttt{<}answer\texttt{>}\texttt{<}CLS\texttt{>}\texttt{<}/answer\texttt{>}. Wrong format or invalid tool call will get a format penalty of $-1$.

\section{Experiments}

To further demonstrate the effectiveness of TAR-FAS, we conduct several additional experiments. We first present quantitative results under the widely used leave-one-out evaluation protocol. We then analyze the behavior of our tool-augmented reasoning model, which is trained under the One-to-Eleven evaluation protocol used in the main paper. Finally, we provide additional reasoning examples across different datasets.

\subsection{ICMO Protocol}

\subsubsection{Implementation Details}

To further evaluate the cross-domain performance of TAR-FAS, we conduct experiments on the widely used four leave-one-out settings and compare its performance with the latest state-of-the-art approaches. This protocol includes the REPLAY-ATTACK (I)~\cite{replayattack}, CASIA-FASD (C)~\cite{casiafasd}, MSU-MFSD (M)~\cite{msumfsd}, and OULU-NPU (O)~\cite{oulunpu} datasets. In each setting, three datasets are used for training (source domains) and the remaining one for testing (target domain). We apply our proposed tool-aware FAS training strategy using ICMO datasets as source domain in both the FAS knowledge transfer stage and the DT-GRPO training stage. All other training hyperparameters remain the same as those described in the implementation details of the main paper.

\subsubsection{Results}

As shown in Table~\ref{tab:p2}, our method achieves a significant performance advantage over all single-modal methods, outperforming them by a substantial margin. This advantage highlights the effectiveness of multimodal learning in enhancing model generalization. Furthermore, our approach also surpasses recent CLIP-based methods, such as FLIP~\cite{flip} and CFPL-FAS~\cite{cfpl}, as evidenced by the average HTER reduction to 1.30\%, compared to 2.98\% and 3.01\%, respectively.
Compared with the previous MLLM-based method I-FAS~\cite{ifas}, our model further improves performance by a small margin, achieving state-of-the-art (SOTA) results. This improvement suggests that incorporating external visual tools enables the model to capture fine-grained visual cues more effectively, thereby enhancing both the robustness and interpretability on FAS task.

\subsection{One-to-Eleven Protocol}

\subsubsection{Fast Answer Performance}

\begin{table}[H]
\small
\centering
\caption{Performance comparison of fast and reasoning answer.}
\label{tab:fast_final}
\begin{tabular}{p{30mm}<{}p{10mm}<{\centering}p{10mm}<{\centering}}
\toprule
\multicolumn{1}{c}{\multirow{2}{*}{}} & \multicolumn{2}{c}{Results}                            \\ \cmidrule(lr){2-3} 
 & HTER(\%) & AUC \\ \midrule
Fast Answer     & 10.54         & 95.06             \\
Reasoning Answer    & \textbf{7.54} & \textbf{96.67}    \\ \bottomrule
\end{tabular}%
\end{table}

To emphasize the effectiveness of tool-augmented reasoning, we compare the performance of the fast answer and the reasoning answer. As shown in \Cref{tab:fast_final}, the reasoning answer outperforms the fast answer with a margin of 1.61\% HTER, demonstrating the effectiveness of incorporating external visual tools.

\subsubsection{Reasoning Accuracy in Training Process}

\begin{figure}[H]
    \centering
    \includegraphics[width=\linewidth]{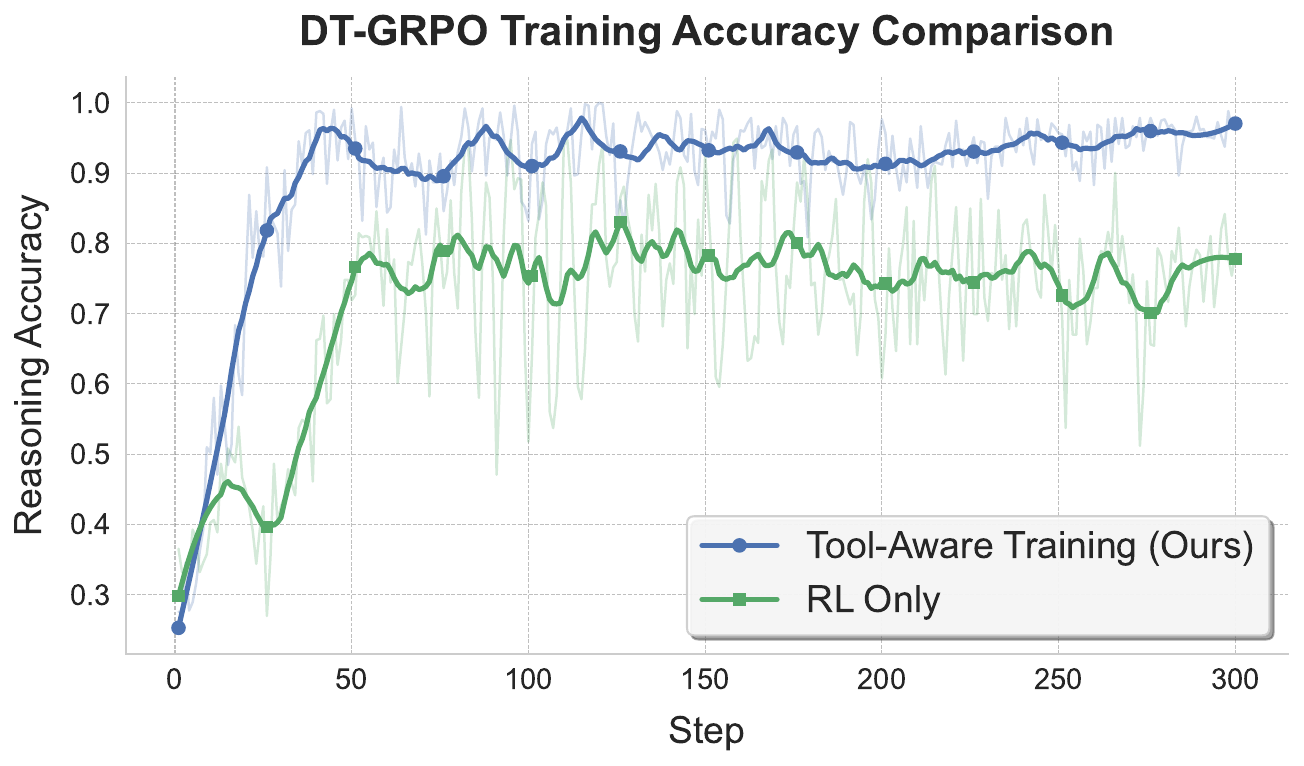}
    \caption{The reasoning accuracy comparison between our \textit{tool-aware training} and \textit{RL only training} during training process.}
    \label{fig:reasoning_accuracy}
\end{figure}

We compare our tool-aware FAS training strategy with a RL only training strategy using the same tool-diversity reward. As shown in \Cref{fig:reasoning_accuracy}, the reasoning accuracy of our method overpass RL only strategy with a substantial margin. The result demonstrate the necessity of knowledge transfer and tool-call format inject process which layes a solid foundation for DT-GRPO.

\subsubsection{Effectiveness of DT-GRPO}

\begin{table}[h]
\small
\centering
\caption{Performance comparison of DT-GRPO and ST-GRPO (Single-Tool GRPO) in DeepEyes \cite{deepeyes}.}
\label{tab:reward}
\begin{tabular}{p{30mm}<{}p{10mm}<{\centering}p{10mm}<{\centering}}
\toprule
\multicolumn{1}{c}{\multirow{2}{*}{RL}} & \multicolumn{2}{c}{Results}                            \\ \cmidrule(lr){2-3} 
 & HTER(\%) & AUC \\ \midrule
ST-GRPO \cite{deepeyes}        & 9.98         & 94.41             \\
\rowcolor[HTML]{C0C0C0} DT-GRPO (Ours)    & \textbf{7.54} & \textbf{96.67}    \\ [-0.6ex] \bottomrule
\end{tabular}%
\end{table}

\noindent To demonstrate the proposed DT-GRPO, we compare the performance against Single-Tool GRPO (ST-GRPO) used in Deepeyes \cite{deepeyes} which enhance multimodal reasoning performance. The reward of ST-GRPO can be formulated as:
\begin{equation}
    R_{ST\text{-}GRPO} = R_{rsn}^{fmt} + R_{rsn}^{acc} + \mathbb{I}_{tool} \cdot \mathbb{I}_{R_{rsn}^{acc} > 0} \,\,\,\, ,
\end{equation}
where $R_{rsn}^{fmt} \in \{ -1, 0 \}$ denotes reasoning format reward, $R_{rsn}^{acc}=\mathbb{I}_{CLS=label}$ denotes reasoning accuracy reward, $\mathbb{I}_{tool} \in \{ 0,1 \}$ indicates whether tool is called. Notably, only ZoomIn tool are allowed in ST-GRPO.

As shown in \Cref{tab:reward}, our proposed DT-GRPO outperforms the ST-GRPO with a substantial margin, gaining generalization performance from diverse tool and tool-diversity reward. Thus demonstrate that our DT-GRPO enables the model to autonomously learn efficient and adaptive diverse tool-use.

\subsubsection{Impact of Maximum Tool-Call Time}

\begin{figure}[H]
    \centering
    \includegraphics[width=\linewidth]{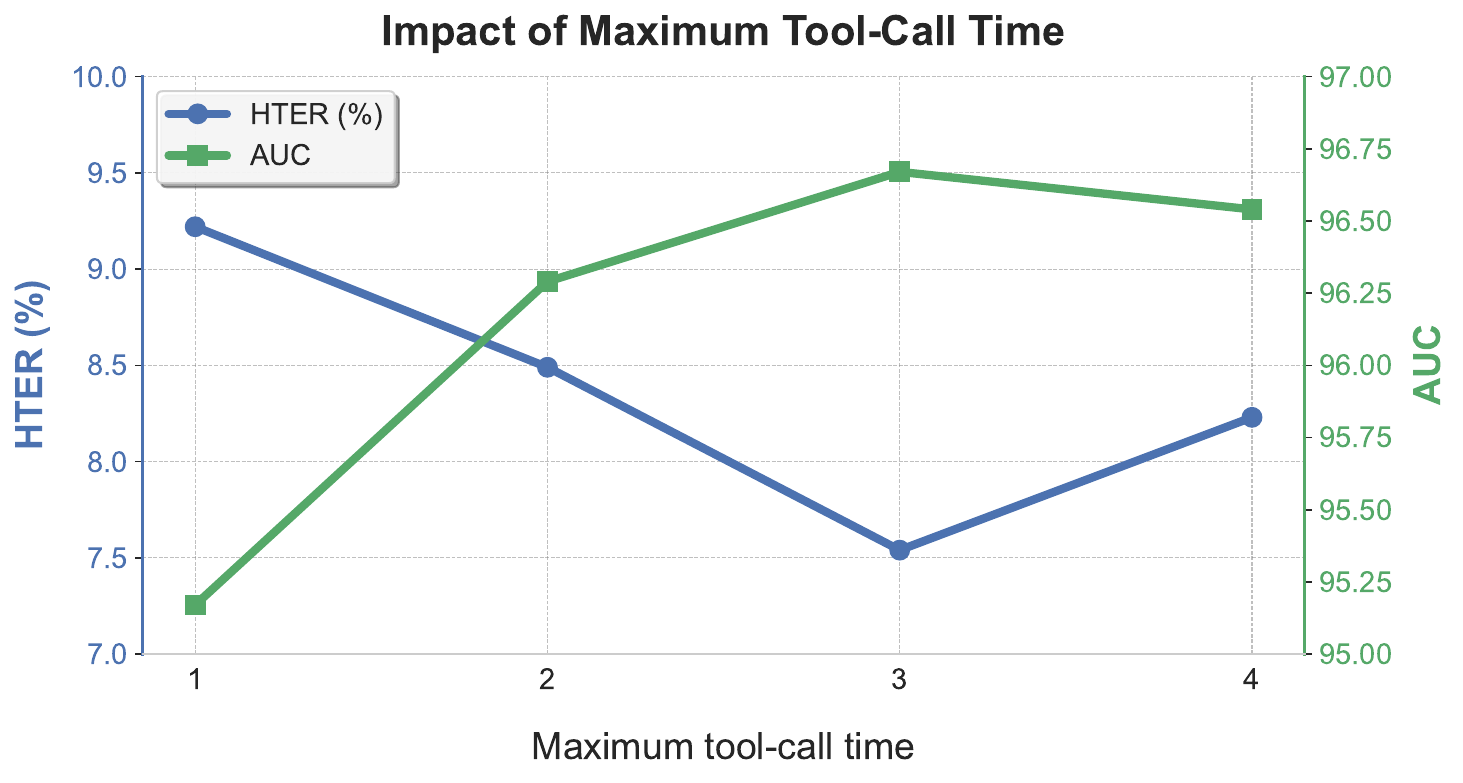}
    \caption{Impact of maximum tool-call time. Increasing the maximum tool-call time from 1 to 3 improves performance, while setting it to 4 leads to a slight drop due to longer rollouts and higher sample uncertainty.}
    \label{fig:tool_call_time_performance}
\end{figure}

We further study the impact of the maximum tool-call limit. During training, this limit is controlled by the maximum number of turns allowed in the rollout process. The relationship between the tool-call limit and model performance is shown in \Cref{fig:tool_call_time_performance}. As illustrated, increasing the limit from 1 to 3 consistently improves performance. However, setting the limit to 4 results in a slight performance drop, likely due to longer rollout and increased sample uncertainty. Therefore, we set the maximum tool-call time to 3 for optimal performance. This experiment further demonstrates the importance of external visual tools in achieving robust FAS performance.

\subsection{Additional Reasoning Samples}

\begin{figure*}[b]
    \centering
    \includegraphics[width=0.8\linewidth]{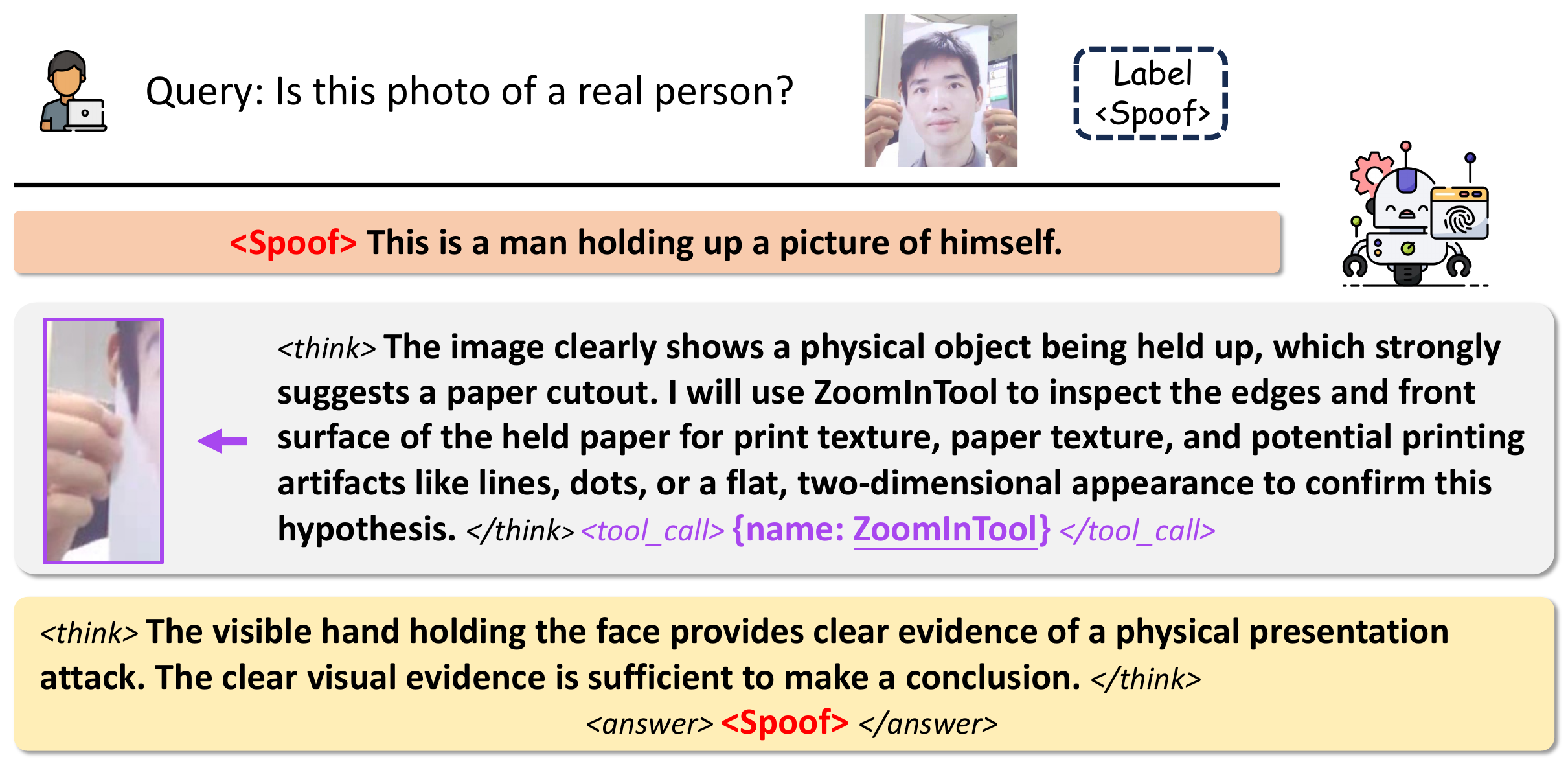}
    \caption{Sample of CASIA-MFSD.}
    \label{fig:visualization_supp_casiamfsd}
\end{figure*}

\begin{figure*}[b]
    \centering
    \includegraphics[width=0.8\linewidth]{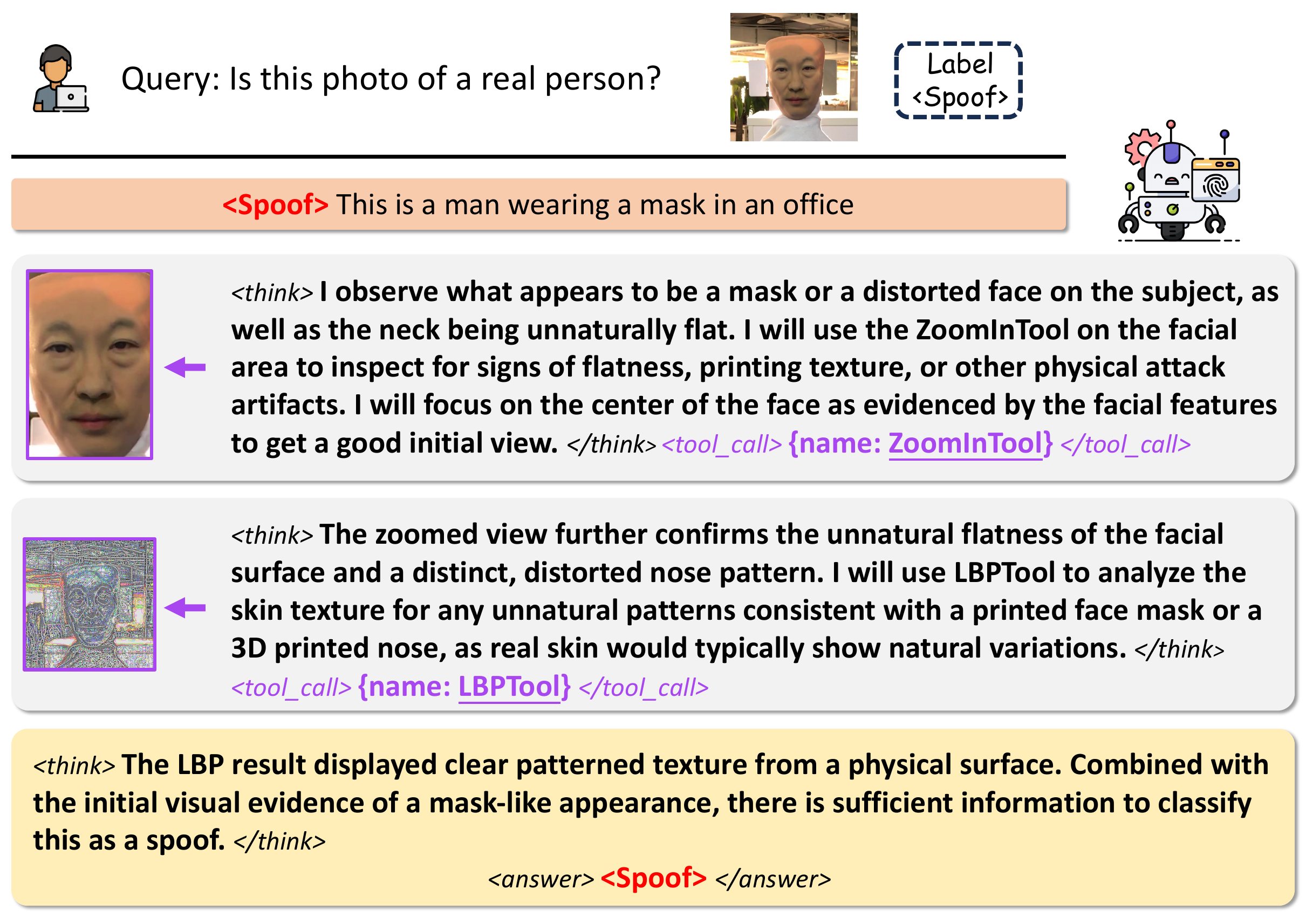}
    \caption{Sample of CASIA-SURF-3DMask.}
    \label{fig:visualization_supp_3dmask}
\end{figure*}

\begin{figure*}
    \centering
    \includegraphics[width=0.8\linewidth]{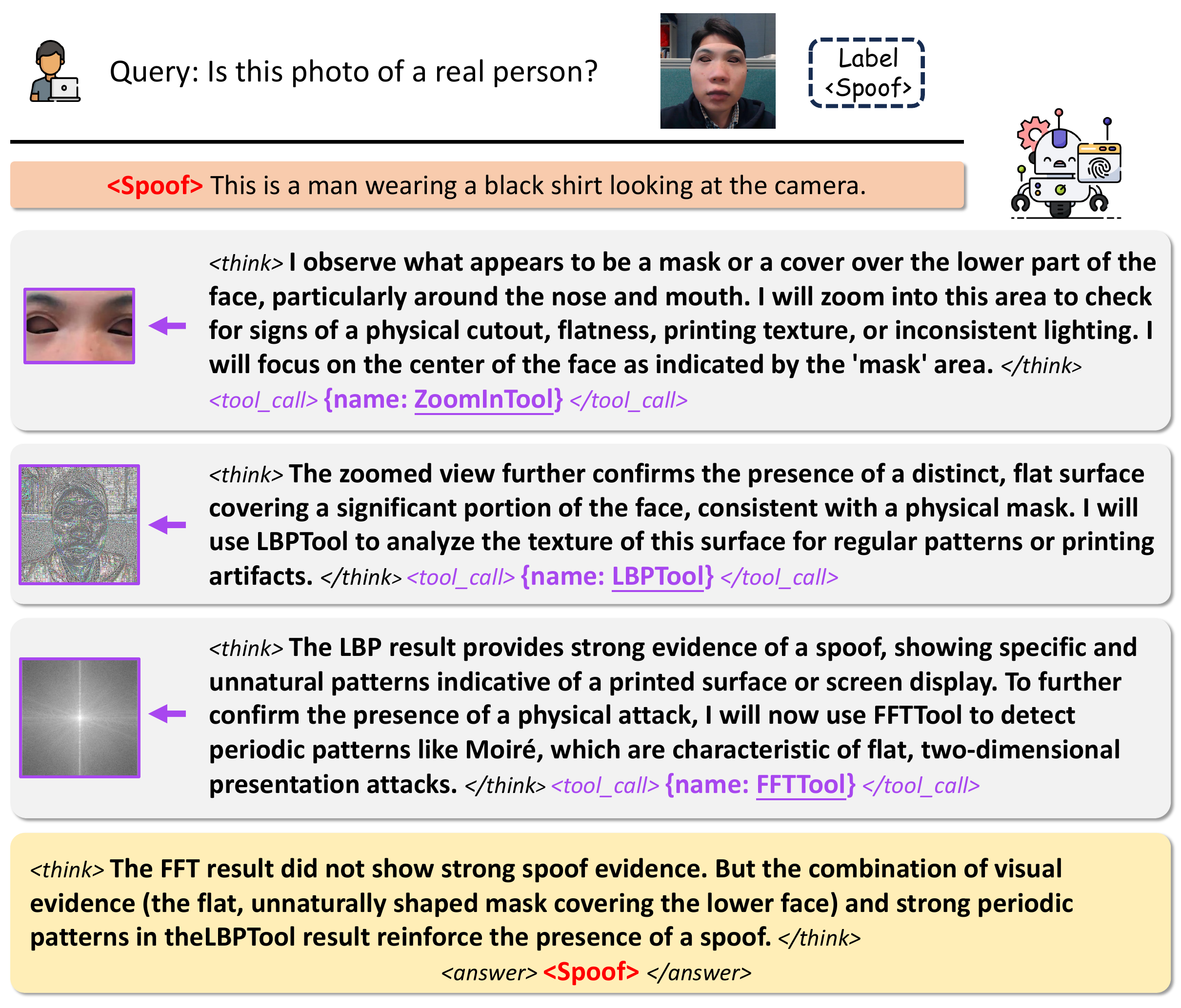}
    \caption{Sample of HKBU-MARs-V1+.}
    \label{fig:visualization_supp_hkbu}
\end{figure*}

\begin{figure*}
    \centering
    \includegraphics[width=0.8\linewidth]{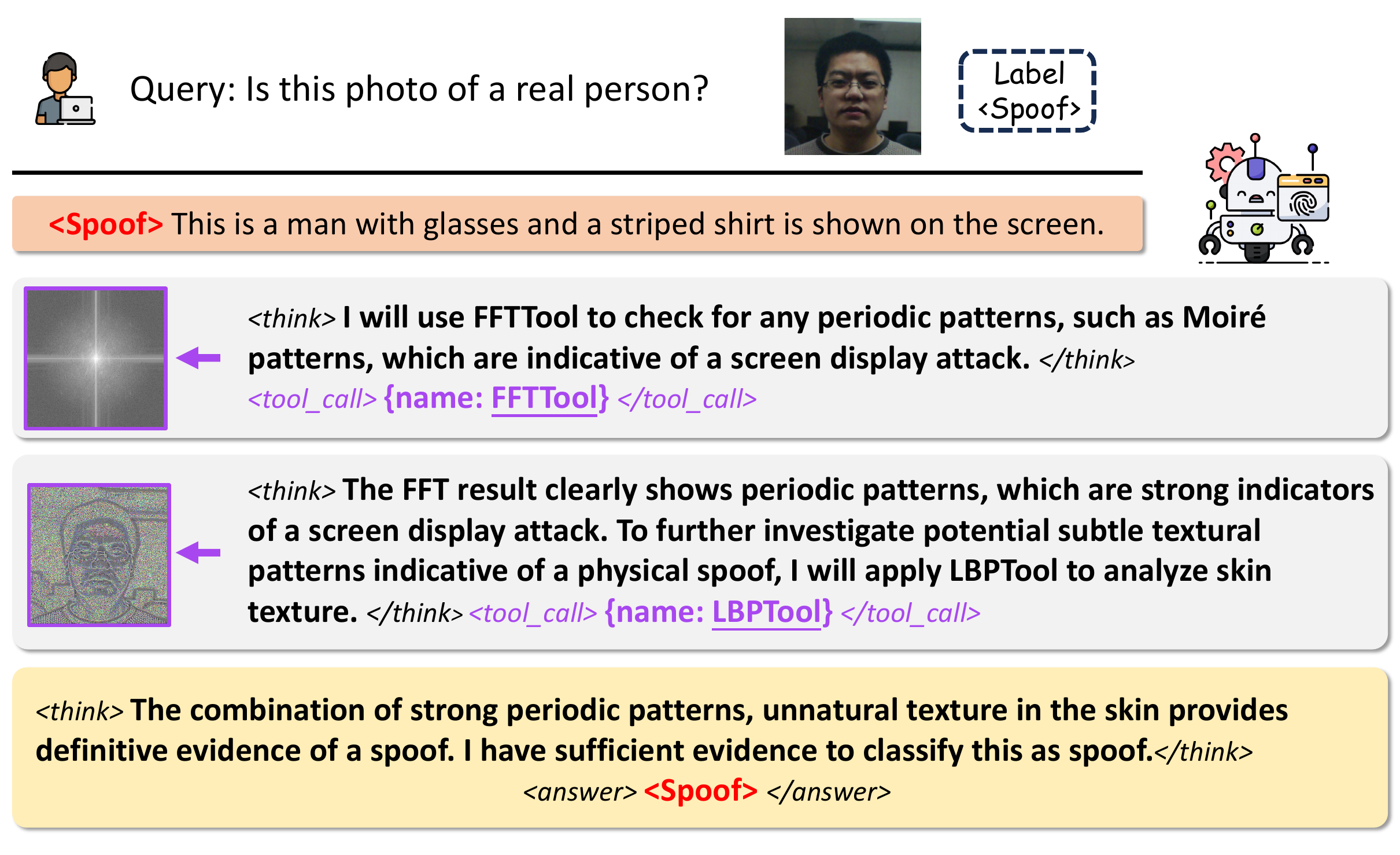}
    \caption{Sample of MSU-MFSD.}
    \label{fig:visualization_supp_msumfsd}
\end{figure*}

\begin{figure*}
    \centering
    \includegraphics[width=0.85\linewidth]{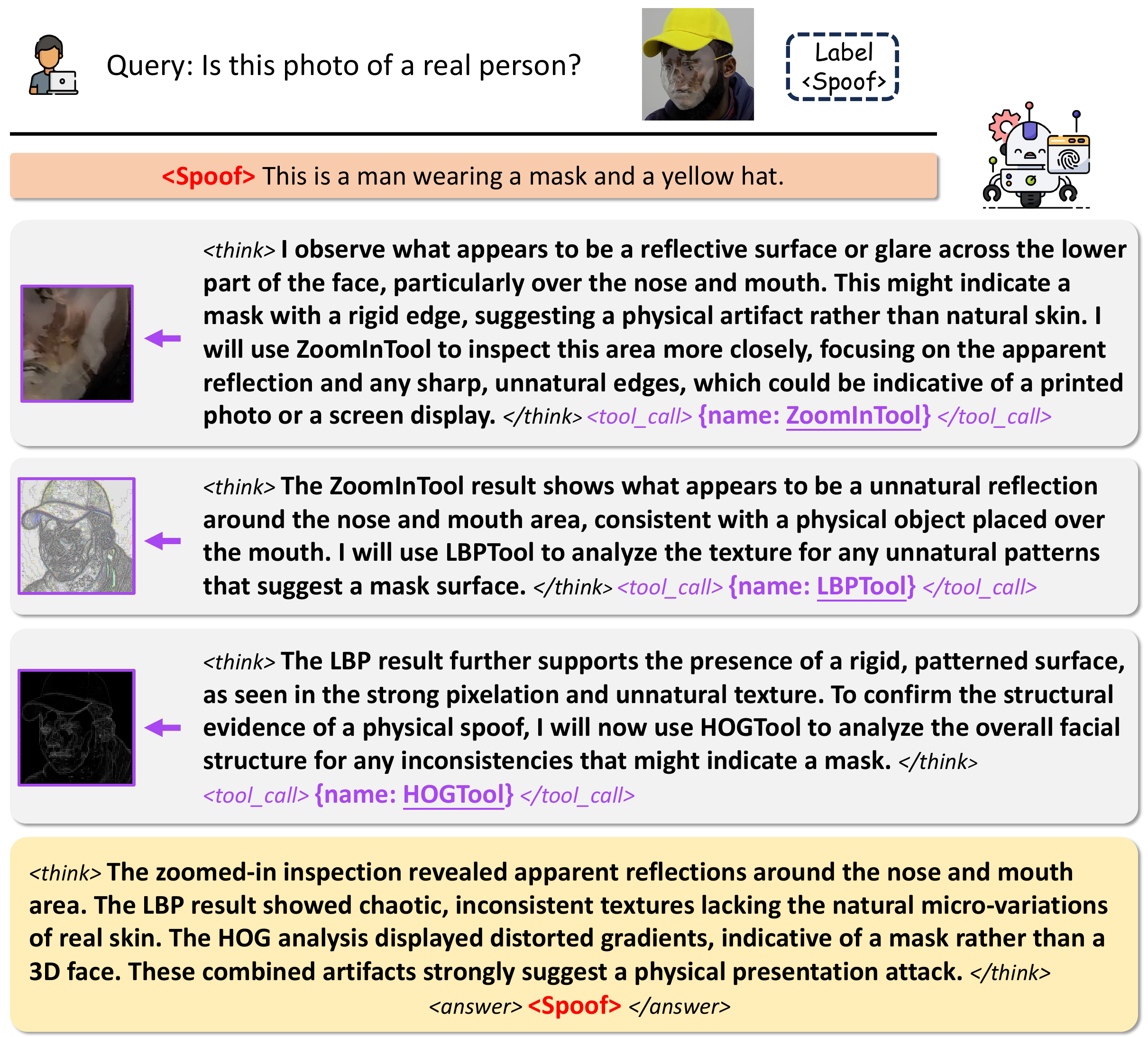}
    \caption{Sample of HiFiMask.}
    \label{fig:visualization_supp_hifimask}
\end{figure*}

\begin{figure*}
    \centering
    \includegraphics[width=0.85\linewidth]{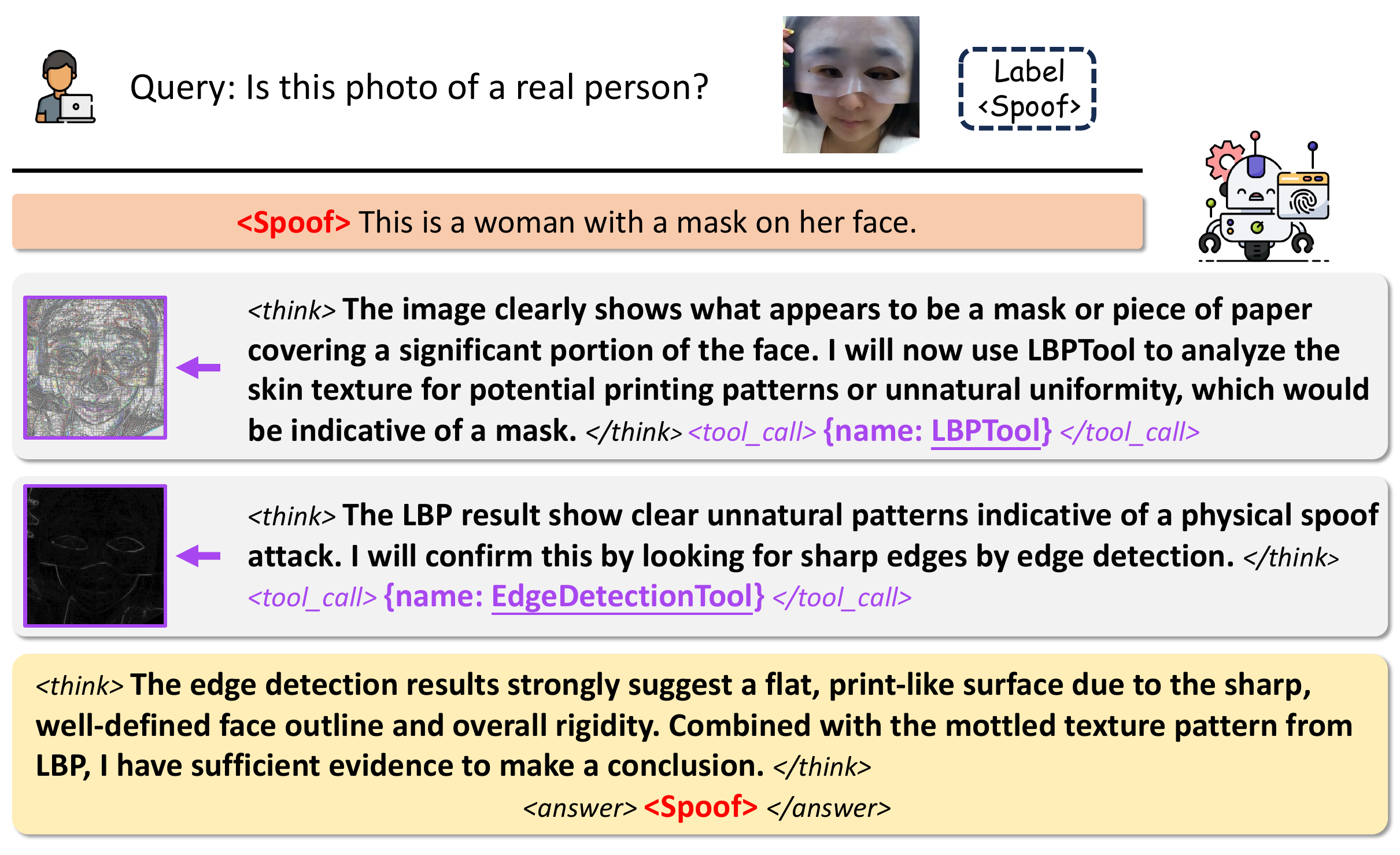}
    \caption{Sample of Rose-Youtu.}
    \label{fig:visualization_supp_roseyoutu}
\end{figure*}

\begin{figure*}
    \centering
    \includegraphics[width=0.85\linewidth]{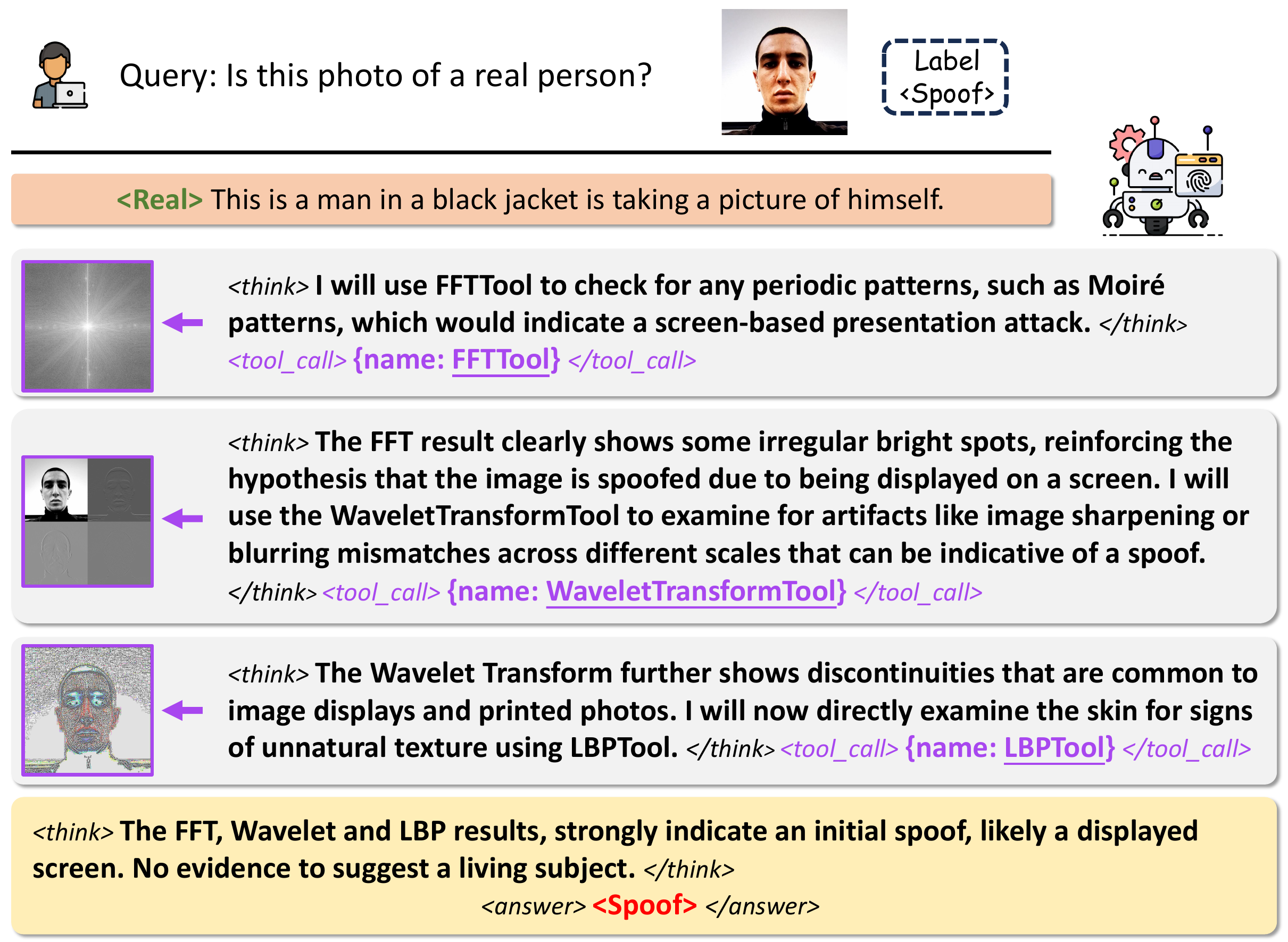}
    \caption{Sample of OULU-NPU.}
    \label{fig:visualization_supp_oulunpu}
\end{figure*}

\begin{figure*}
    \centering
    \includegraphics[width=0.85\linewidth]{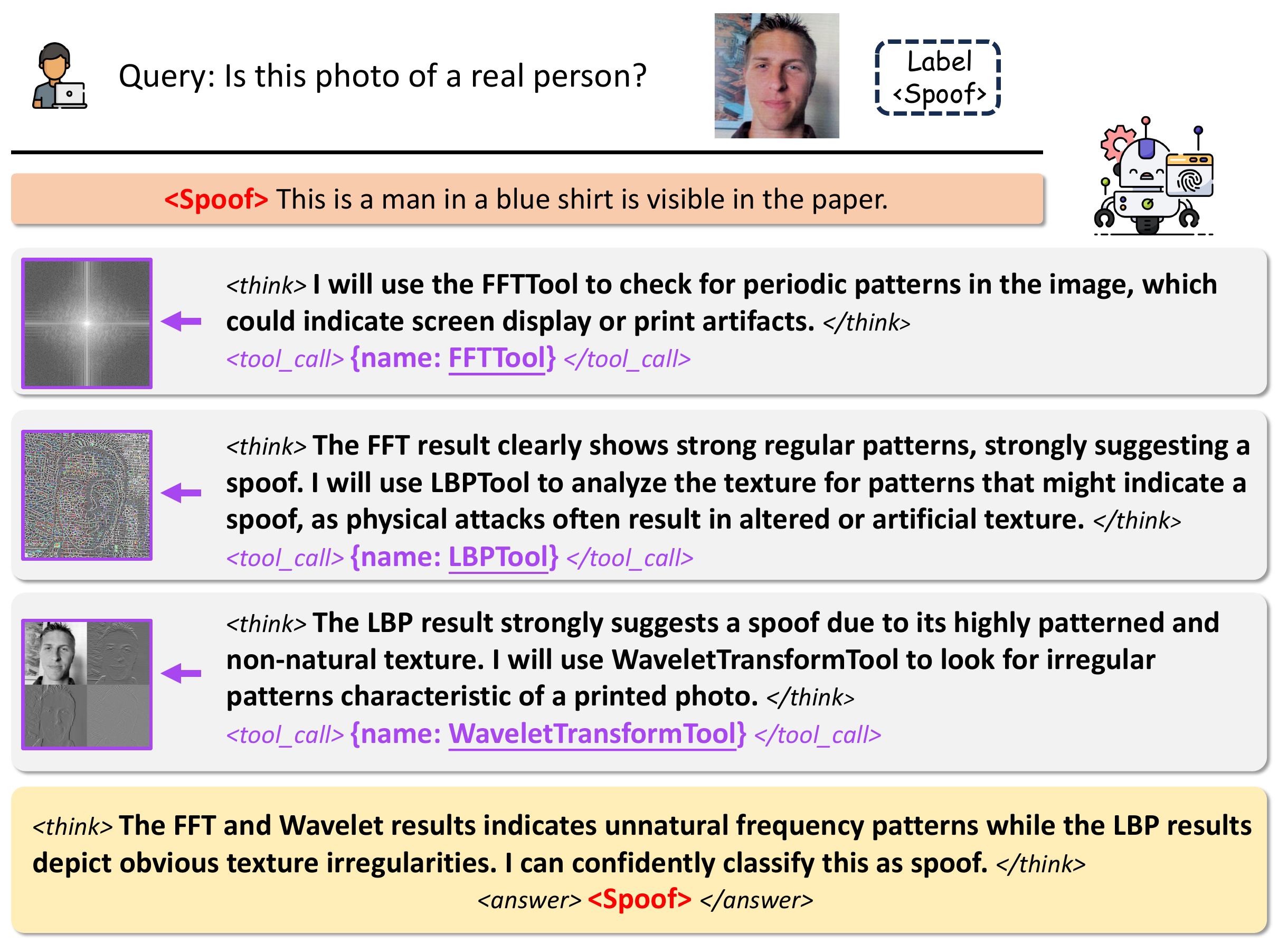}
    \caption{Sample of REPLAY-ATTACK.}
    \label{fig:visualization_supp_reaplyattack}
\end{figure*}

\begin{figure*}
    \centering
    \includegraphics[width=0.85\linewidth]{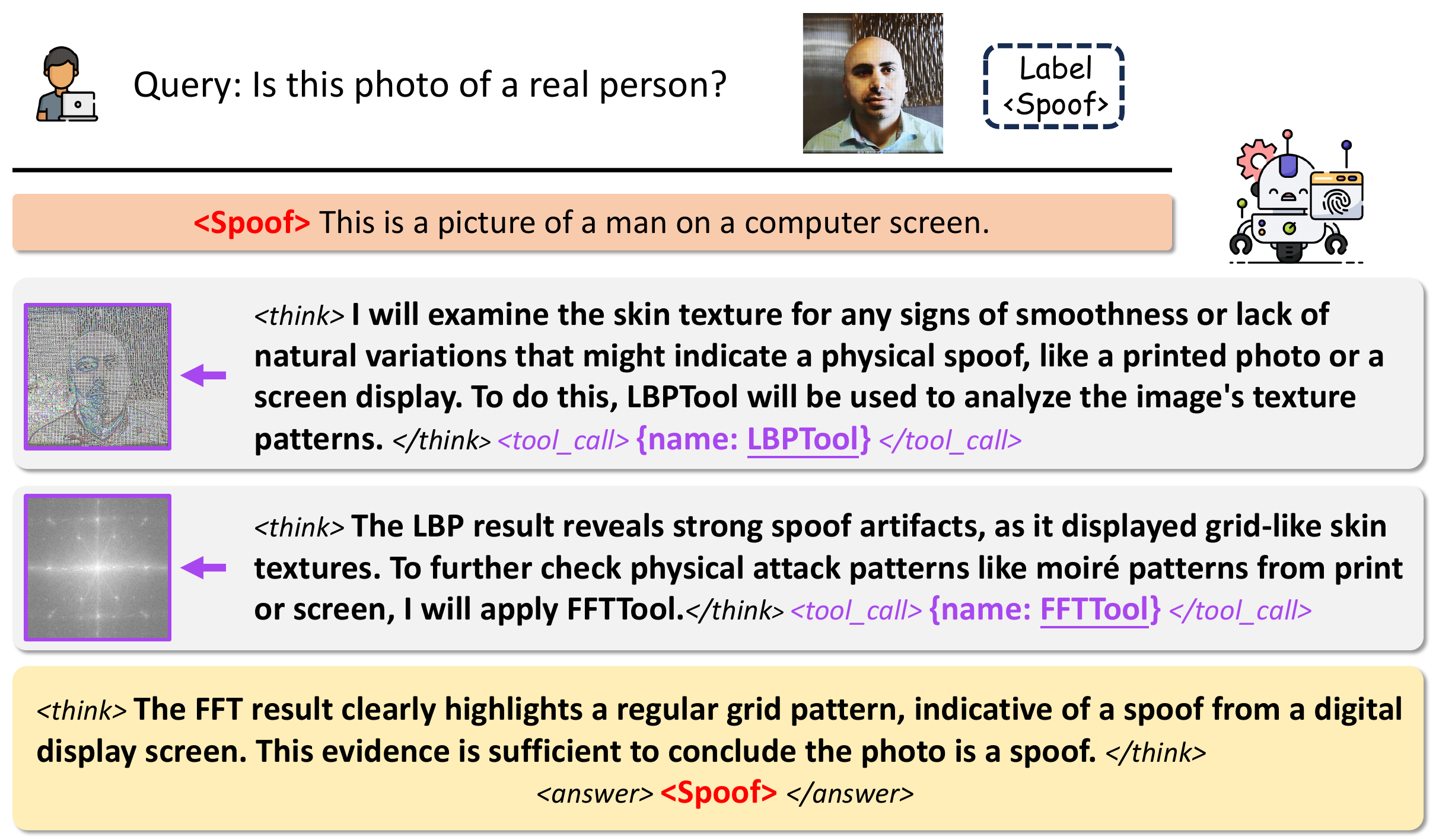}
    \caption{Sample of SIW.}
    \label{fig:visualization_supp_siw}
\end{figure*}

\begin{figure*}
    \centering
    \includegraphics[width=0.85\linewidth]{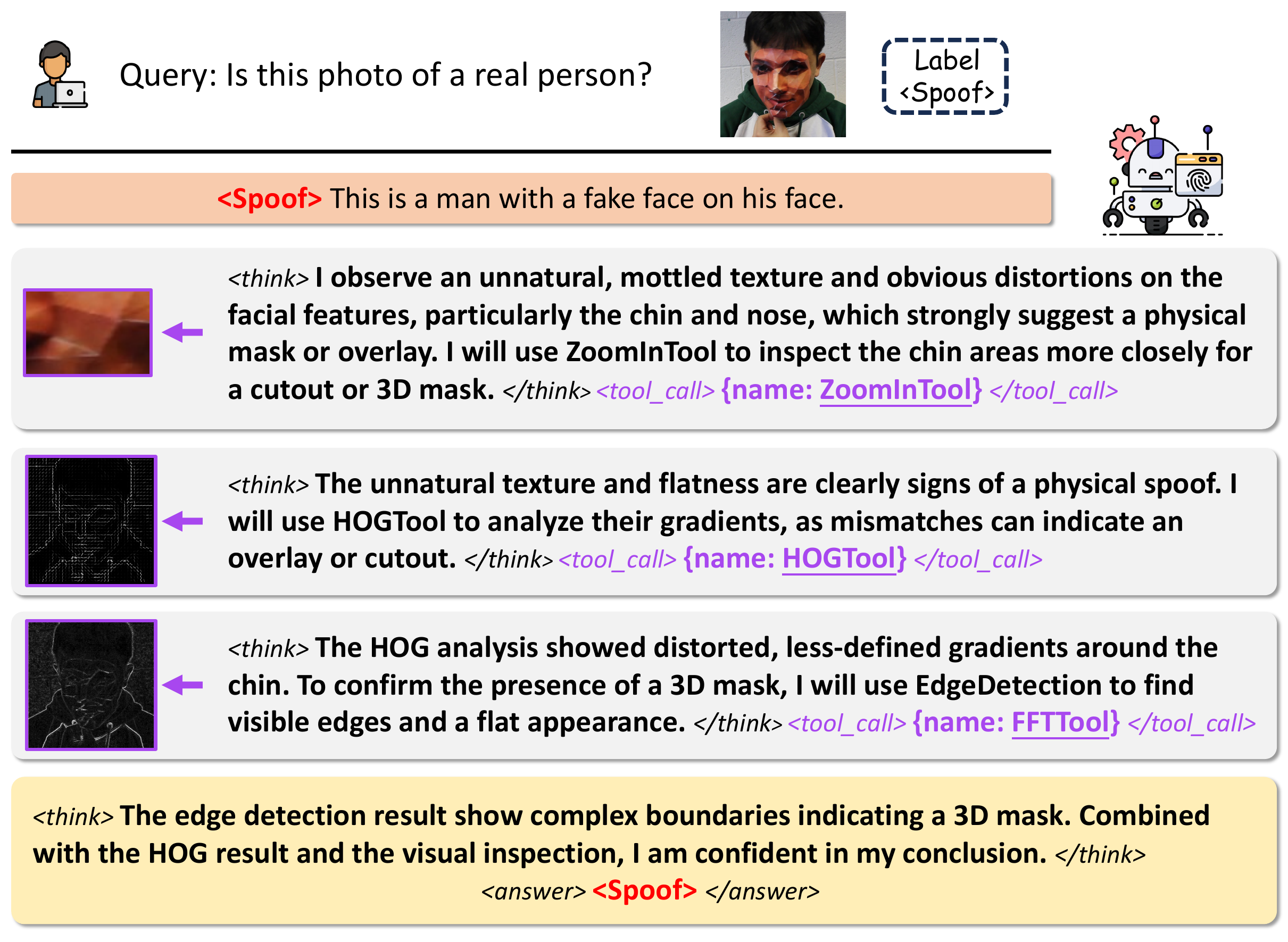}
    \caption{Sample of SIW-M-V2.}
    \label{fig:visualization_supp_siwmv2}
\end{figure*}

\begin{figure*}
    \centering
    \includegraphics[width=0.85\linewidth]{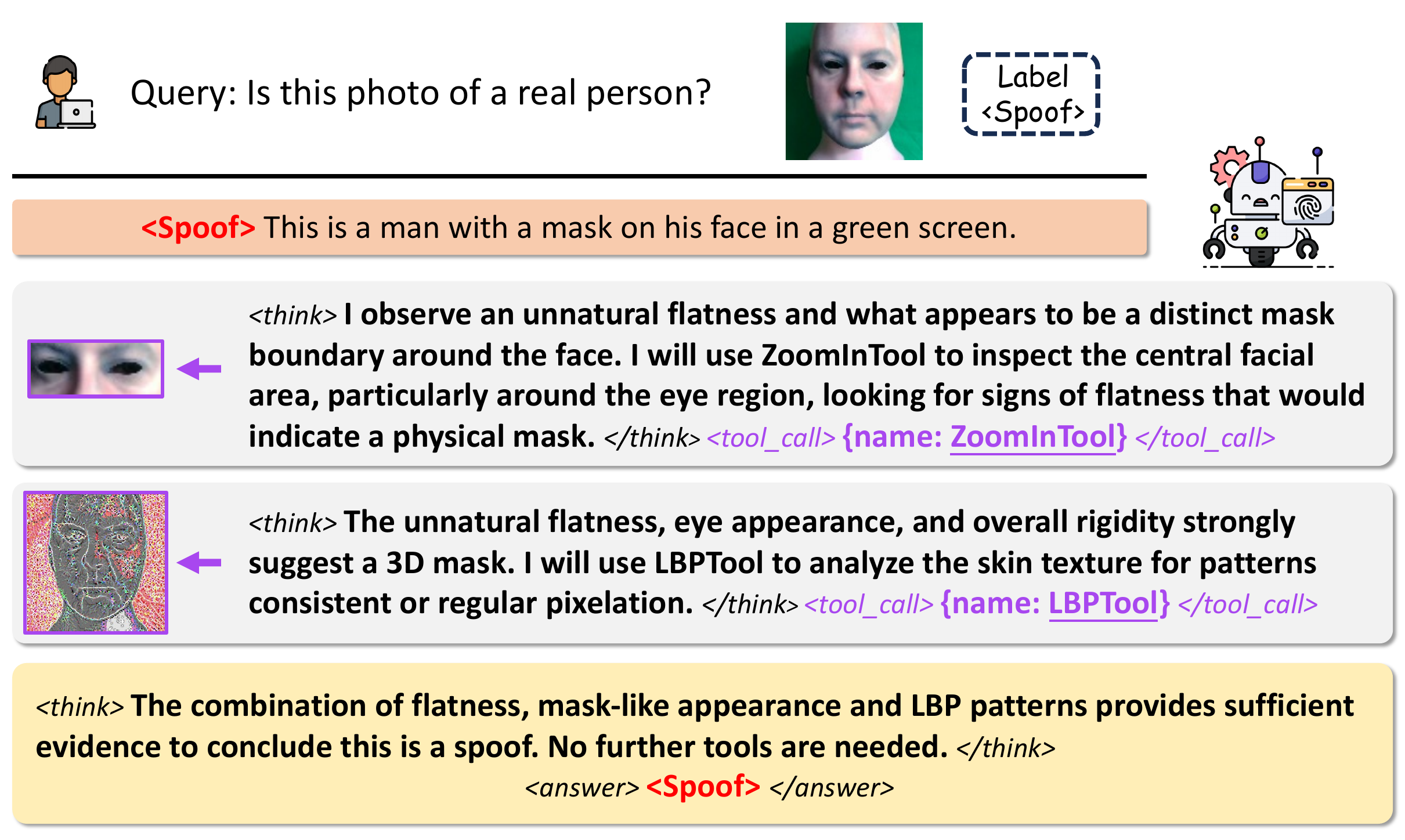}
    \caption{Sample of WMCA.}
    \label{fig:visualization_supp_wmca}
\end{figure*}

To further exhibit the tool-augmented reasoning process of TAR-FAS, we choose one sample from each of the eleven evaluated datasets. The results are shown in \Cref{fig:visualization_supp_casiamfsd}-\ref{fig:visualization_supp_wmca}.

\FloatBarrier
\clearpage

\end{document}